\documentclass{article}

\usepackage{microtype}
\usepackage{graphicx}
\usepackage{subcaption}
\usepackage{booktabs} 

\usepackage{url}
\usepackage[ruled,linesnumbered,vlined]{algorithm2e}
\usepackage{subcaption}
\usepackage{xcolor}
\usepackage{multirow}

\usepackage{hyperref}

\usepackage[preprint]{icml2026}

\usepackage{amsmath}
\usepackage{amssymb}
\usepackage{mathtools}
\usepackage{amsthm}

\usepackage[capitalize,noabbrev]{cleveref}

\theoremstyle{plain}

\theoremstyle{definition}

\theoremstyle{remark}

\usepackage[textsize=tiny]{todonotes}

\icmltitlerunning{InfoLaw: Information Scaling Laws for Large Language Models with Quality-Weighted Mixture Data and Repetition}

\begin{document}

\twocolumn[
  \icmltitle{InfoLaw: Information Scaling Laws for Large Language Models with Quality-Weighted Mixture Data and Repetition}



  \icmlsetsymbol{equal}{*}

  \begin{icmlauthorlist}
    \icmlauthor{Fengze Liu}{equal,bty}
    \icmlauthor{Weidong Zhou}{equal,bty}
    \icmlauthor{Binbin Liu}{bty}
    \icmlauthor{Ping Guo}{bty}
    \icmlauthor{Zijun Wang}{bty,ucsc}
    \icmlauthor{Bingni Zhang}{bty}
    \icmlauthor{Yifan Zhang}{bty}
    \icmlauthor{Yifeng Yu}{bty}
    \icmlauthor{Xiaohuan Zhou}{bty}
    \icmlauthor{Taifeng Wang}{bty}
  \end{icmlauthorlist}

  \icmlaffiliation{bty}{ByteDance}
  \icmlaffiliation{ucsc}{UC Santa Cruz}

  \icmlcorrespondingauthor{Fengze Liu}{fengze.liu@bytedance.com}
  \icmlcorrespondingauthor{Weidong Zhou}{zhouweidong.66@bytedance.com}

  \icmlkeywords{Machine Learning, ICML}

  \vskip 0.3in
]



\printAffiliationsAndNotice{}  

\begin{abstract}

  Upweighting high-quality data in LLM pretraining often improves performance, but in data-limited regimes, especially under overtraining, stronger upweighting increases repetition and can degrade performance. However, standard scaling laws do not reliably extrapolate across mixture recipes or under repetitions, making the selection for optimal data recipes at scaling underdetermined. To solve this, we introduce \textbf{InfoLaw} (Information Scaling Laws), a data-aware scaling framework that predicts loss from consumed tokens, model size, data mixture weights, and repetition. The key idea is to model pretraining as information accumulation, where quality controls information density and repetition induces scale-dependent diminishing returns. We first collect the model performance after training on datasets that vary in scale, quality distribution, and repetition level. Then we build up the modeling for information so that information accurately predicts those model performance. InfoLaw predicts performance on unseen data recipes and larger-scale runs (up to 7B, 425B tokens) with 0.15\% mean and 0.96\% max absolute error in loss, and it extrapolates reliably across overtraining levels, enabling efficient data-recipe selection under varying compute budgets.
  
\end{abstract}

\section{Introduction}
Training large language models (LLMs) requires access to high-quality data \citep{Brown:2023,Chowdhery:2023}. However, the availability of high-quality data is severely limited \citep{Villalobos_24}, and in the data-constrained settings, upweighting higher-quality data inevitably increases repetition, which has been shown to impair performance when excessive \citep{Muennighoff_2023}. This issue is further exacerbated by the widespread adoption of overtraining \citep{llama,qwen3}—a strategy that reduces inference costs compared to the compute-optimal regime \citep{chinchila}. 

\begin{figure*}
    \centering
    \includegraphics[width=1.0\textwidth]{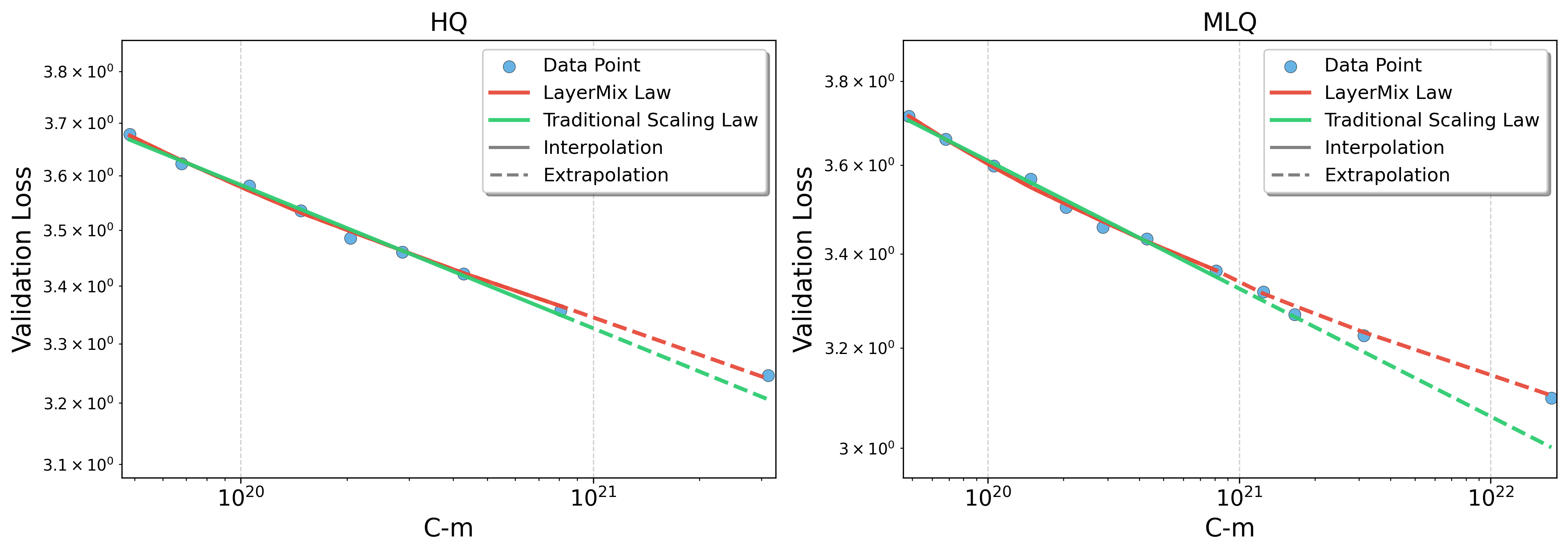}
    \caption{Validation loss versus compute $C_m$ in the loss--$C$ view under LayerMix data with repetition. Curves are fit on 252M--1.2B and extrapolated to larger models. The traditional scaling law mis-extrapolates under repetition, while InfoLaw tracks both interpolation and extrapolation across recipes (HQ and MLQ).}
    \label{fig:info_vs_scalinglaw}
\end{figure*}

To address the shortage of high-quality data as model scale increases, a common compromise is to incorporate lower-quality data, thereby reducing the repetition of high-quality samples. Intuitively, high-quality data provides greater performance gains than low-quality data upon first exposure, but as repetition increases, the marginal benefit decays—eventually approaching that of unseen low-quality data. However, the optimal balance between quality and repetition remains unclear. A standard approach for identifying optimal mixing strategies is to run smaller-scale experiments and extrapolate performance to larger compute budgets using scaling laws \citep{gpt4,chinchila,Chowdhery:2023}. Yet, as shown in Figure \ref{fig:info_vs_scalinglaw}, under conditions of data repetition, standard scaling laws fail to reliably predict model performance at scale \citep{Danny_2022,Muennighoff_2023}. Moreover, they do not generalize across different mixing strategies , necessitating grid searches over data recipes—an approach that is costly even at small scales.

In this paper, we study the problem of scaling large language models in a data-aware regime, where training data consists of a heterogeneous mixture with varying quality levels, and each quality level is repeated to different extents. We introduce a theoretical framework, the InfoLaw, which accounts for both the scaling effects of mixture weights and the impact of repetition. Our formulation views training as a process of accumulating information from the dataset, with model performance determined by the total information gained by the end of training. At each step, the information gain is modeled as the sum of contributions from different quality ranges. Within each quality range, the gain depends on two factors: an information density function, parameterized by quality (with higher quality assigned higher density), and an exponential decay term that captures the interactions between model scale, data scale, and repetition level.

To fit the parameters of the InfoLaw, we construct a suite of datasets that vary along three axes: scale, quality, and repetition level. Specifically, we partition the source dataset into buckets according to quality scores, and then sample from each bucket with different weights, a procedure we refer to as LayerMix sampling. Following the data-constrained setting, the source dataset is first downsampled to the target scale to ensure stable repetition effects. We then train 9 models ranging from 252M to 1.2B parameters from scratch, each under the same 3.6x over-trained ratio \citep{overtrainscalinglaw}. For each model, we construct three datasets with distinct LayerMix sampling configurations, resulting in 27 total training runs. Model performance is evaluated as the average perplexity across five downstream tasks. Finally, we fit the InfoLaw to these results, estimating the parameters that best capture the relationship between information gain and observed performance.

We evaluate the generalization of InfoLaw along three axes: (i) unseen mixture recipes (new LayerMix sampling weights), (ii) larger compute scales, and (iii) a higher overtraining ratio ($25\times$). Across these settings, InfoLaw accurately predicts loss on unseen recipes and scales up to a 7B model trained on 425B tokens, with 0.15\% mean and 0.96\% maximum absolute error. Moreover, using the fitted law we search over candidate mixtures and identify a data recipe for a 2.5B model that outperforms four randomly sampled baselines without additional training runs. The same parameters also extrapolate well to the $25\times$ overtraining regime.

\section{Related Work}

\paragraph{Scaling Laws} Empirical studies have shown that transformer language models exhibit predictable power-law scaling with model size and training data \citep{EmpiricalSL,Attention,Chowdhery:2023,gpt2}, which has motivated the development of many large-scale systems, including dense models \citep{gpt3,Gopher,Llama3} and mixture-of-experts variants \citep{DeepSeek_V3,qwen3,moe1}. Compute-based scaling laws further formalize how to allocate model capacity and training tokens under a fixed compute budget: \citet{chinchila} characterized the compute-optimal regime, while subsequent work explored alternative allocations and the interaction between compute $C$ and optimization choices such as batch size and learning rate \citep{openaisl,deepseeksl}.

In parallel, training smaller models on substantially more tokens than the compute-optimal point has become increasingly common for efficiency and deployment reasons \citep{llama,qwen3}. \citet{BeyondChinchilla} extended the Chinchilla framework by incorporating factors such as data quality and inference requirements, and \citet{overtrainscalinglaw} showed that scaling relations can remain reliable in overtrained regimes. For predicting downstream performance, \citet{ScalingDownStream} studied how downstream metrics scale after fine-tuning, and \citet{Mirage} linked non-linear evaluation metrics to perplexity, supporting perplexity as a more stable proxy than earlier observations of emergent/unstable metrics \citep{Emergent}.

\paragraph{Data-Aware Scaling} Traditional scaling laws often assume effectively unlimited data, but in practice high-quality data is scarce and therefore frequently upsampled \citep{Multilingual}. Under repetition, prior work reports diminishing returns and, beyond some point, performance degradation when upsampling subsets or repeating datasets \citep{Danny_2022,Muennighoff_2023}. At the same time, \citet{xue2023repeat} suggests that, in certain regimes, continuing to train on repeated data can still be preferable to stopping early, highlighting that the effect of repetition is non-trivial and not captured by classical laws. More recently, \citet{subscalinglaw} studied how scaling interacts with data density, providing a finer-grained view in limited-data regimes.

A separate line of work uses scaling laws to optimize data recipes. \citet{datamixinglaw} incorporated mixture weights into loss prediction, and \citet{autoscale} argued that optimal mixing can be model-scale dependent. \citet{RegMix} uses proxy models to search mixture ratios without training the full-scale model, while \citet{CMR,DCPT} leverage scaling insights in continued pre-training and domain-mixture design; \citet{ScalingQuality} further analyzes the interaction between scaling and data quality. In contrast, our goal is to predict loss under quality-weighted mixtures \emph{with explicit repetition}, enabling extrapolation across both mixture recipes and repetition levels.

\section{Limitations of Conventional Scaling Laws}

In this section, we reveal and substantiate a critical limitation of conventional scaling laws in the context of data repetition and quality selection. First, we introduce the LayerMix sampling function in section \ref{sub:layermix}, to imitate real scenario where the data is a mixture of different quality and repetition degrees. Next, we compare the relationship between the model's loss $L$ and amount of compute $C$ in cases with and without repetition in section \ref{sub:loss-C}, and the results show that the traditional scaling law performs well on data without repetition

\subsection{LayerMix Sampling Function}
\label{sub:layermix}

\paragraph{Source Data} We obtain our training corpora from Common Crawl \citep{commoncrawl}, following \citet{RefinedWeb} and obtain 15T English tokens. We ran global fuzzy deduplication across all snapshots to ensure there is no repeat data in the corpora. The final dataset contains 3.7T token. Details are in Appendix \ref{app:train-set}.

\begin{table*}
    \caption{Preset LayerMix sampling weights and Searched optimal sampling weights for 2.5B model. }
    \label{tab:layermix_param}
    \centering
    \begin{tabular}{l|cccccc}
        \hline
        \textbf{Name} & \textbf{$w0$} & \textbf{$w1$} & \textbf{$w2$} & \textbf{$w3$} & \textbf{$w4$} & \textbf{$w5$}\\  
        \hline
        \textbf{HQ  (High Quality)} & 0.80 & 0.10 & 0.03 & 0.03 & 0.02 & 0.0 \\
        \textbf{MHQ  (Medium-High Quality)} & 0.66 & 0.22 & 0.05 & 0.03 & 0.02 & 0.0 \\
        \textbf{MQ (Medium Quality)} & 0.48 & 0.23 & 0.13 & 0.07 & 0.07 & 0.0 \\
        \textbf{MLQ (Medium-Low Quality)} & 0.38 & 0.21 & 0.20 & 0.11 & 0.08 & 0.0 \\
        \textbf{LQ  (Low Quality)} & 0.24 & 0.20 & 0.19 & 0.18 & 0.17 & 0.0 \\
        \textbf{Optimal Recipe of 2.5B model with $m=3.6$} & 0.50 & 0.49 & 0.01 & 0.0 & 0.0 & 0.0 \\
        \hline
    \end{tabular}
\end{table*}

\paragraph{Training Data Sampling}
We assign each document a quality score following \citet{quadmix}: we apply two quality classifiers
\citep{finewebedu,dclm} and take the average of their normalized scores. We rank all documents by this
score and partition the corpus into six buckets by percentile: 0--5\%, 5--20\%, 20--40\%, 40--60\%,
60--80\%, and 80--100\%.

We then define a LayerMix sampling function $H(w, K, S, B)$ to construct a packed training set.
Here $S$ is the number of tokens in the source corpus to sample from, $K$ is the total number of tokens
in the packed training set (we use one-epoch training to avoid additional epoch-induced repetition),
$w=[w_0,\dots,w_5]$ with $\sum_d w_d=1$ specifies the target token proportions of the six buckets in the
training set, and $B=[B_0,\dots,B_5]$ specifies the bucket proportions in the source corpus (in our
setting $B=[0.05,0.15,0.20,0.20,0.20,0.20]$).

For bucket $d$, the training set contains $K_d=w_dK$ tokens sampled from $S_d=B_dS$ source tokens.
Let $M_d=\min(K_d,S_d)$ denote the number of {unique (non-repeated) tokens} from bucket $d$ that
appear in the packed training set, and define the average repetition factor as $R_d=K_d/M_d=w_dK/M_d$,
so $R_d=1$ when $K_d\le S_d$ and $R_d>1$ otherwise. The full packing procedure is given in
Appendix~\ref{app:layermix}.

By varying $(w,K,S)$, LayerMix produces datasets with different scale, quality mixture, and repetition.
We enforce $w_d \ge w_{d+1}$ to keep higher-quality buckets more represented. We use five preset
mixtures (HQ, MHQ, MQ, MLQ, LQ; Table~\ref{tab:layermix_param}), and set $w_5=0$ to drop the lowest
20\% bucket. Unless stated otherwise, we set $K=S$ to isolate the repetition effects induced by $w$.

\subsection{Traditional Scaling Law Between Loss and Amount of Compute}
\label{sub:loss-C}

We compare the relationship between model loss $L$ and total compute $C$ under regimes with and without repetition in an overtrained setting. Specifically, under the compute-optimal scheme, $C_{opt}=N_{opt} K_{opt}$, where $K$ is the consumed tokens, $N$ is the non-embedding FLOPs per token as defined in \citet{deepseeksl} and $N_{opt}$, $K_{opt}$ is the Chinchilla-optimal pair. Then in the overtrained setting, following \citet{overtrainscalinglaw}, we set $K_m=\sqrt{m}K_{opt}$, $N_m=\frac{1}{\sqrt{m}}N_{opt}$, $C_m=K_mN_m$ with $m=3.6$. And \citet{overtrainscalinglaw} shows that the the Loss–Compute relation preserves the fitted exponent for models trained with the same overtraining factor $m$.




The model loss is collected by training on datasets sampled with LayerMix parameters HQ and MLQ, see details in Table \ref{tab:layermix_param}. Dataset HQ has more high quality data but with more repitition, while MLQ has more diverse data with less repitition. We then visualize the relationship between compute $C_m$ and model loss $L$ in the loss--$C_m$ view in Figure \ref{fig:info_vs_scalinglaw}. Here $L$ is the average perplexity over five downstream tasks—HellaSwag~\citep{hellaswag}, ARC-E/ARC-C~\citep{ARC}, MMLU~\citep{MMLU}, and TriviaQA~\citep{triviaqa}. Following~\citet{Mirage}, we convert downstream accuracies into perplexity to obtain a smoother scaling signal. As shown in Figure~\ref{fig:info_vs_scalinglaw}, although a conventional power-law scaling curve can interpolate within the fitting regime (252M--1.2B), it systematically mis-extrapolates as $C_m$ increases under LayerMix data with repetition, yielding overly optimistic loss reductions. This failure appears consistently across representative mixture recipes, indicating that compute alone is insufficient to characterize scaling behavior in the presence of quality-weighted mixtures and repetition.

These observations suggest that traditional scaling laws are not reliably predictive under quality-weighted mixture data with repetition, especially for extrapolation. Therefore, we need a modified scaling law that explicitly incorporates both the data quality distribution and the degree of data repetition as core variables.

\section{Information Scaling Laws}

In this section, we introduce the design of InfoLaw. We treat the training process as gaining information from the dataset and propose to calculate Information as accumulation of information gain throughout the training process, which synthesizes the impacts of data quality, repetition level, model scales and total training tokens, and then build power-law relationship with the model's final validation loss.

\subsection{Information Measurement}
\label{sub:info}

To build intuition for how repetition interacts with data quality, we compare two 850M runs trained with different LayerMix sampling weights. In the more repetition-heavy recipe (HQ), the top 5\% quality bucket is repeated by roughly $16\times$, whereas in a less repetitive recipe (MQ) it is repeated by roughly $10\times$. Empirically, the two runs achieve similar evaluation loss early in training, but the more repetitive run improves substantially more slowly in the later stage and converges to a worse final loss, indicating diminishing returns from repeated exposures. See Appendix~\ref{app:motivation_figs} 
 Figure~\ref{fig:app_motivation}(b) for the training-time curves.


\begin{figure*}
    \centering
    \begin{subfigure}[t]{0.45\textwidth}
        \centering
        \includegraphics[width=\textwidth]{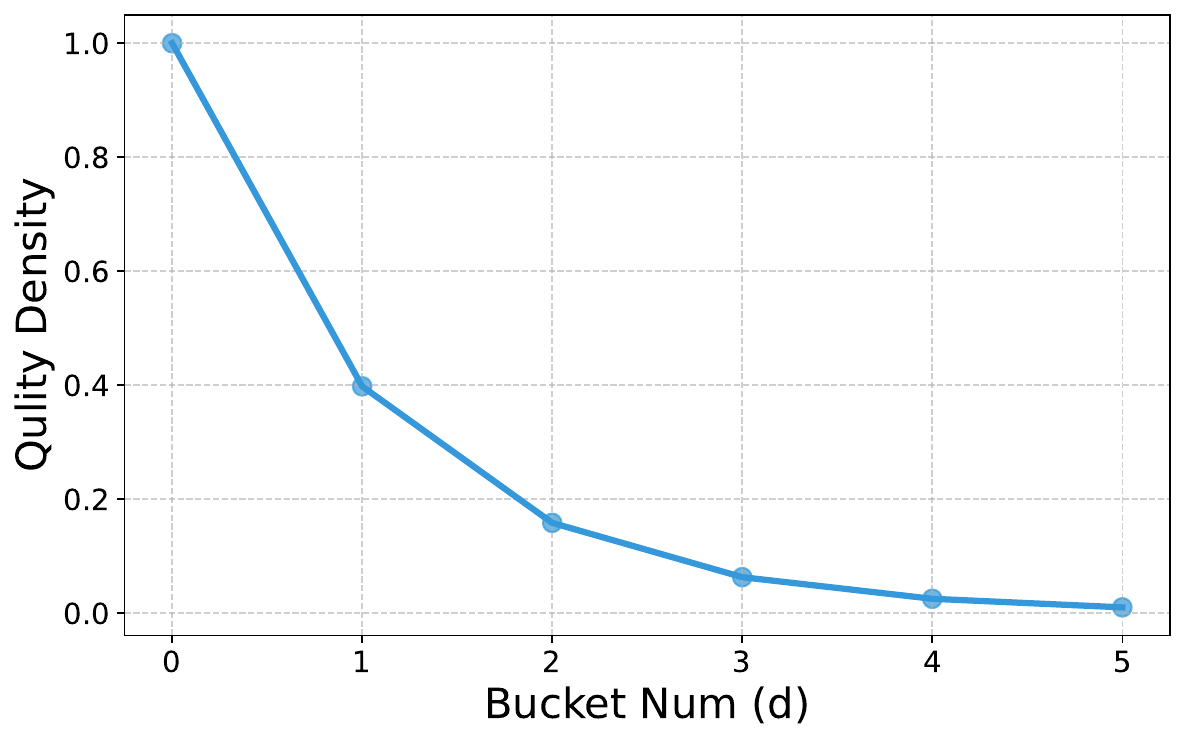}
        \caption{The fitted quality density function $f_d$. The quality density is a monotonically decreasing function of the bucket index, meaning buckets with higher-quality data are assigned a higher density value.}
        \label{fig:quality_density}
    \end{subfigure}
    \hfill
    \begin{subfigure}[t]{0.45\textwidth}
        \centering
        \includegraphics[width=\textwidth]{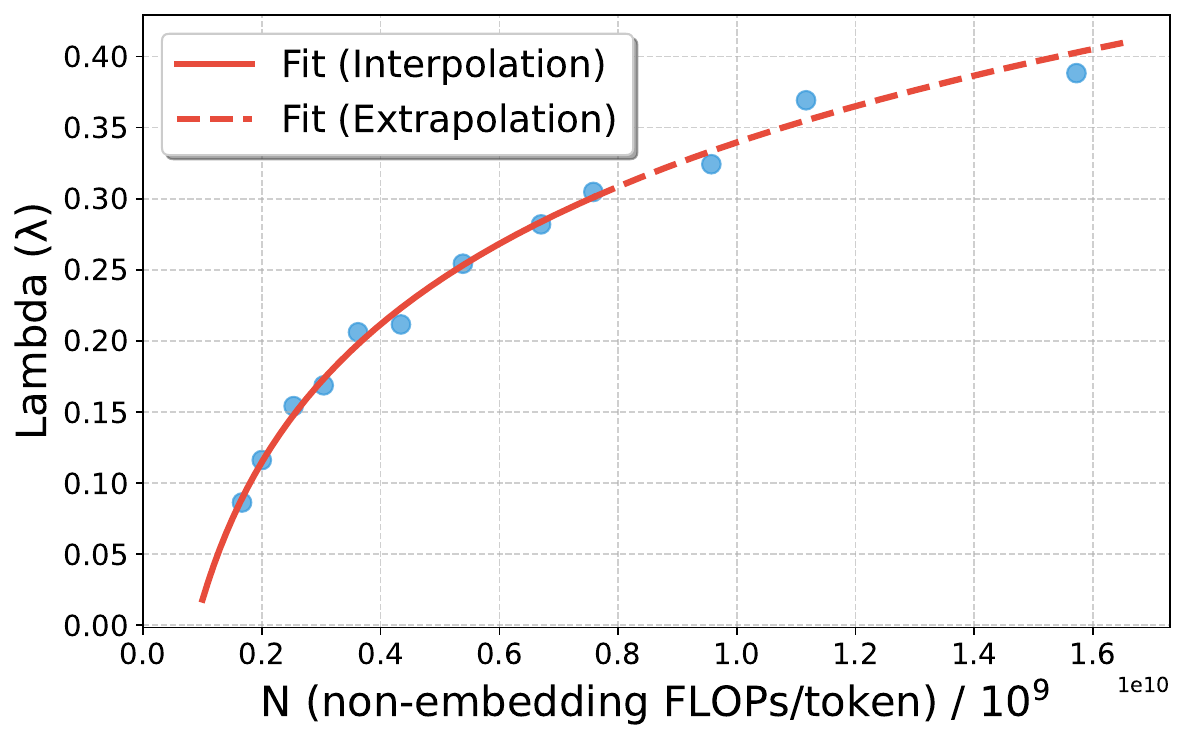}
        \caption{The relationship between $\lambda$ and $N$ with a fitted curve. The blue scattered points represent the observed data. The solid red line shows the fit within the data range, while the dashed line represents the extrapolation.}
        \label{fig:lmbd-N}
    \end{subfigure}
    
     \caption{The fitted function of quality density function and relationship between $\lambda(N)$ and $N$ }
\end{figure*}

Based on this observation, we propose an exponential decay function to model the decreasing information gain of repeated data. Assuming the Information a document $i$ contains is $I_i$, then the information a language model gets at $t$-th learning from the document $i$ is:

\begin{equation}
I_{i\_\text{part}}(t, \lambda(N)) = I_i \cdot \lambda(N) e^{-\lambda(N) t}
\label{eq:I_part}
\end{equation}

where $\lambda(N)$ is a nonnegative rate parameter that depends on the model’s non-embedding FLOPs/token $N$ and is fitted from data.

When a language model learning the document for total $T$ times, the Information learned from the document is:

\begin{equation}
I_{i\_\text{total}}(T, \lambda(N)) = \int_0^T I_{i\_\text{part}}(t, \lambda(N))dt = I_i\cdot(1-e^{-\lambda(N) T})
\label{eq:I_total}
\end{equation}

Equation \ref{eq:I_total} captures the principle of diminishing returns in learning: repeated exposure to a document yields progressively smaller gains, causing the total acquired information to saturate and asymptotically approach the document's full information content $I_i$.

To capture the empirically observed slowdown in marginal gains relative to the total training budget $K$, we incorporate a logarithmic normalization factor. This formulation is empirically grounded and essential for generalizing the scaling law across orders of magnitude in training volume, as validated in Appendix \ref{app:normalization_justification}.

\begin{equation}
I_{i\_\text{part}}(t, \lambda(N), K) = I_i \cdot \lambda(N)e^{-\lambda(N) t / \log(K)}
\label{eq:K_part}
\end{equation}

Then the Equation \ref{eq:I_total} becomes to:

\begin{equation}
\label{eq:K_total}
\begin{aligned}
I_{i, \text{total}} & (t, \lambda(N), K) = \int_0^T I_{i, \text{part}}(t, \lambda(N), K) dt \\
&= I_i \cdot \log(K) \left( 1 - e^{-\lambda(N) T / \log(K)} \right)
\end{aligned}
\end{equation}

For all the training data, we sum them together as the final Information the language model learned from the training corpora, denoting as $\text{info}$:

\begin{equation}
\label{eq:info}
\begin{aligned}
& \text{info}(w, K, S, f, \lambda(N)) \\
& \quad = \sum_d I_d \cdot \log(K) \left( 1 - e^{-\lambda(N) R_d / \log(K)} \right) \\
& \quad = \sum_d f_d M_d \log(K) \cdot \left( 1 - e^{-\lambda(N) R_d / \log(K)} \right)
\end{aligned}
\end{equation}

where $d$ is the quality bucket number from 0 to 5. $I_d$ is the total Information in d-the bucket, which can be calculated by the multiplication of number of unique tokens $M_d=min(w_dK,B_dS)$ and information density $f_d$, which is a parameterized quality density function. $R_d=\frac{w_dK}{M_d}$ is the average repeat times for the data from the $d$-th bucket and $\lambda(N)$ is related with $N$, which are to be fitted from the data.

Equation \ref{eq:info} can be divided into two parts: the first term is $I_d=f_dM_d  \log(K)$, it represents the total Information contained in the packed data of the $d$-th bucket, and the second term is $1-e^{-\lambda(N) R_d/\log(K)}$, it represents the language model's learning ability on this data when repeated an average of $R_d$ times. And the total Information learned by the language model is the product of these two terms.

\begin{figure*}[t]
  \centering

  \begin{subfigure}{0.24\textwidth}
    \includegraphics[width=\linewidth]{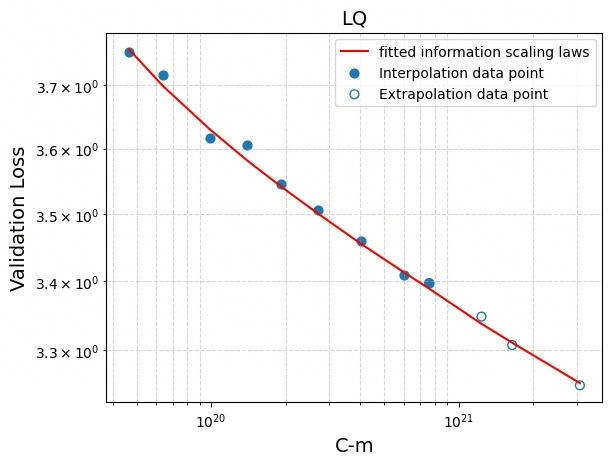}
    \caption{}
  \end{subfigure}
  \hfill
  \begin{subfigure}{0.24\textwidth}
    \includegraphics[width=\linewidth]{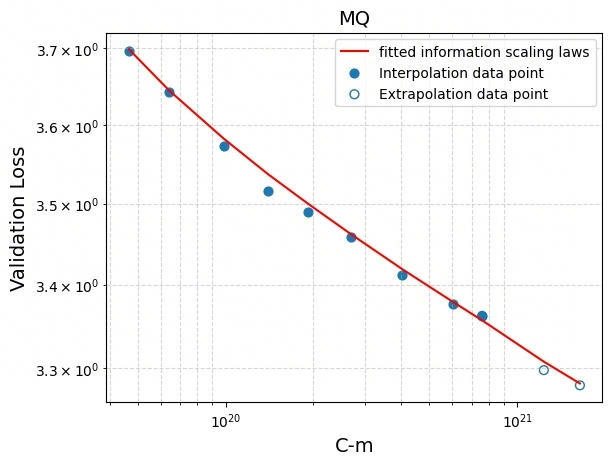}
    \caption{}
  \end{subfigure}
  \hfill
  \begin{subfigure}{0.24\textwidth}
    \includegraphics[width=\linewidth]{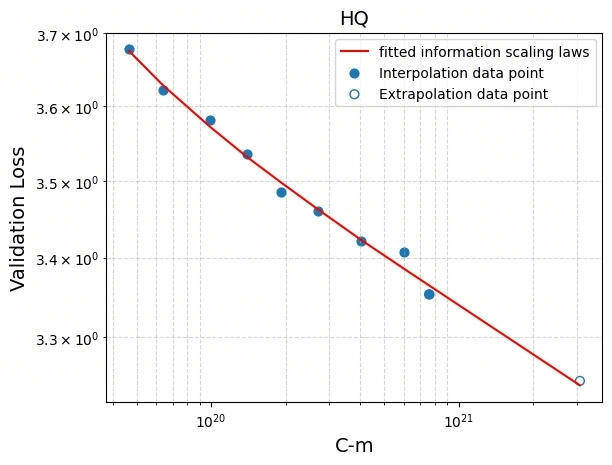}
    \caption{}
  \end{subfigure}
  \begin{subfigure}{0.24\textwidth}
    \includegraphics[width=\linewidth]{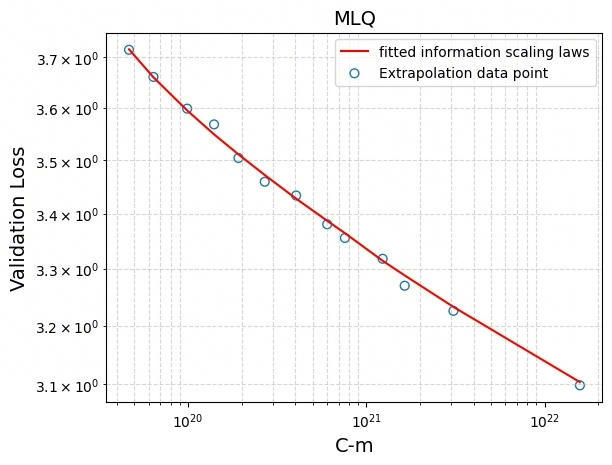}
    \caption{}
  \end{subfigure}

  \vspace{0.5em}

  \begin{subfigure}{0.24\textwidth}
    \includegraphics[width=\linewidth]{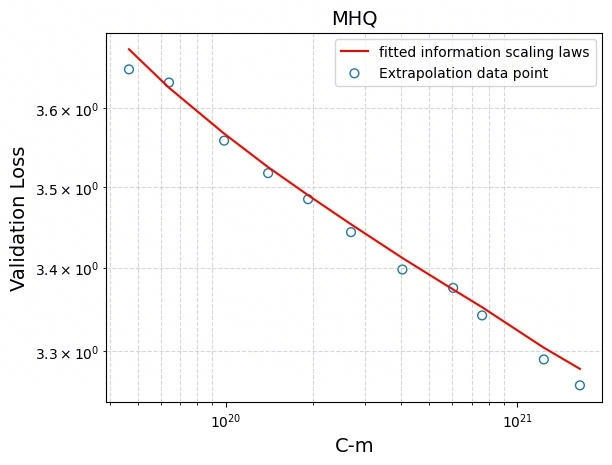}
    \caption{}
  \end{subfigure}
  \hfill
  \begin{subfigure}{0.24\textwidth}
    \includegraphics[width=\linewidth]{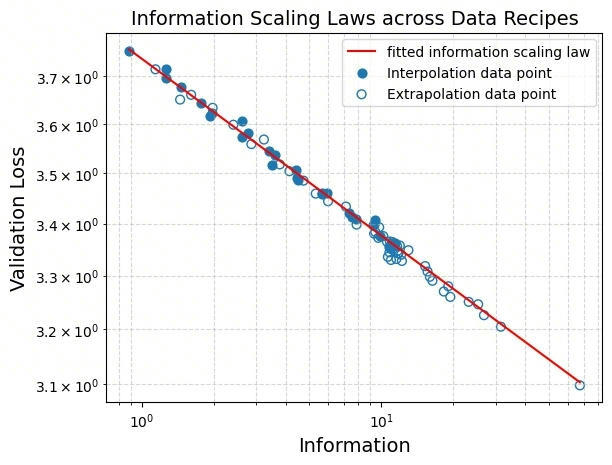}
    \caption{}
    \label{fig:infolaw}
  \end{subfigure}
  \hfill
  \begin{subfigure}{0.24\textwidth}
    \includegraphics[width=\linewidth]{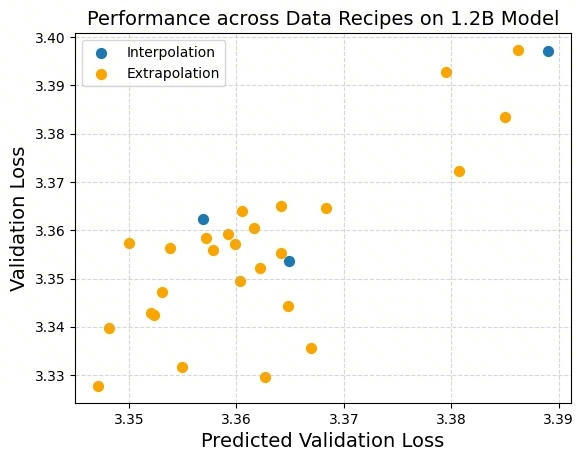}
    \caption{}
  \end{subfigure}
  \hfill
  \begin{subfigure}{0.24\textwidth}
    \includegraphics[width=\linewidth]{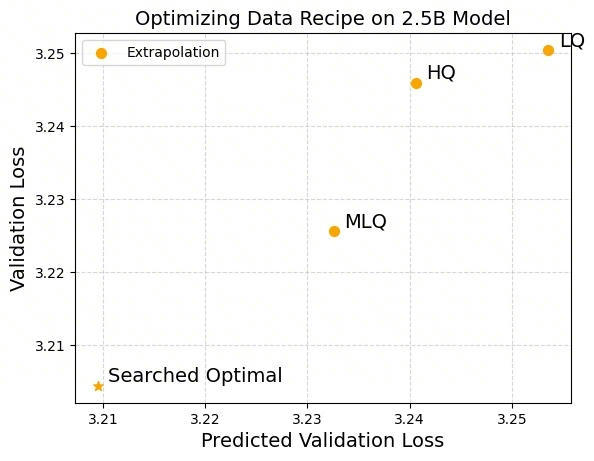}
    \caption{}
    \label{fig:result_2B5}
  \end{subfigure}

  \caption{\textbf{Verification, Unification, and Application of Information Scaling Laws.} 
    Panels \textbf{(a)-(e)} demonstrate that information scaling laws hold independently across varying data quality distributions (LQ to MHQ), consistently following power-law trajectories. 
    \textbf{(f)} Illustrates the \textbf{Information Scaling Laws}, where diverse data recipes collapse onto a single curve when mapped to the information quantity metric, confirming the universality of the law.
    \textbf{(g)} Validates predictive capability on a 1.2B model, showing strong correlation between predicted and actual validation loss for both interpolation and extrapolation settings. 
    \textbf{(h)} Demonstrates optimization on a 2.5B model, where the "Searched Optimal" recipe identified by our framework achieves lower validation loss compared to fixed baselines.}
    \label{fig:scaling_laws_all}
  \label{fig:seven_panels}
\end{figure*}

We propose Information, a metric computed from LayerMix sampling weights $w$, train token $K$ and two fitted functions ($f_d$, $\lambda(N)$, to quantify the knowledge learned during training. Since it is designed to be monotonic with model performance, it enables loss prediction for various training configurations prior to any actual runs. The fitting of $f_d$ and $\lambda(N)$ is described in Section \ref{sec:fit-curve}.

\subsection{Information-Loss Power Law}
\label{sub:loss-info}

As illustrated in Figure~\ref{fig:info_vs_scalinglaw}, in the loss--$C_m$ view conventional scaling laws are not reliably predictive under quality-weighted mixture data with repetition, with extrapolation errors that grow at larger compute. This motivates replacing compute with a repetition and quality aware effective data signal. In the next section, we show that our Information collapses results across mixture recipes and scales onto a unified power-law curve.

We use the Information proposed in Section \ref{sub:info} and plot the $L$-$info$ figure. As illustrated in Figure \ref{fig:infolaw}, when we replace the traditional computation axis $C$ with our novel metric: Information, the experimental points with different LayerMix sampling weights $w$, Model non-embedding FLOPs/token $N$ and Train Token $K$ now collapse perfectly onto a single, unified power-law curve, where they were previously scattered and separated. 

Then the relationship between the loss $L$ and $info$ can be measured using power-law formulation as:

\begin{equation}
L = \alpha \cdot info^{-\beta}
\label{eq:loss-C}
\end{equation}
    
In our experiment, $\alpha=3.7373$ and $\beta=0.0441$. We show them in a log-log plot, so it appears as a straight line with a slope of $-\beta$ and an intercept of $\log(\alpha)$. 

Like the traditional scaling law \citep{chinchila}, we can now conduct experiments on small models to compare the advantages and disadvantages of different experimental configurations, and then use our proposed information scaling law to extrapolate the performance of larger models under larger training tokens.

\begin{table*}
\centering
\caption{The best data recipe for different models and train token}
\label{tab:best_mix}
\begin{tabular}{cccccccccc}
\hline
\textbf{Model} & \textbf{Train Token} & \textbf{Source Token} & \textbf{$w0$} & \textbf{$w1$} & \textbf{$w2$} & \textbf{$w3$} & \textbf{$w4$} & \textbf{$w5$} \\
\hline
\multirow{3}{*}{7B}
& 300B & 500B & 0.548 & 0.444 & 0.004 & 0.003 & 0.002 & 0.000 \\
& 500B & 500B & 0.496 & 0.492 & 0.007 & 0.003 & 0.002 & 0.000 \\
& 800B & 500B & 0.439 & 0.430 & 0.130 & 0.001 & 0.000 & 0.000 \\
& 1000B & 500B & 0.395 & 0.387 & 0.214 & 0.003 & 0.001 & 0.000 \\
\hline
\multirow{3}{*}{1.8B}
& 300B & 500B & 0.619 & 0.376 & 0.004 & 0.001 & 0.000 & 0.000 \\
& 500B & 500B & 0.548 & 0.444 & 0.004 & 0.003 & 0.002 & 0.000 \\
& 800B & 500B & 0.496 & 0.492 & 0.007 & 0.003 & 0.002 & 0.000 \\
& 1000B & 500B & 0.491 & 0.487 & 0.017 & 0.005 & 0.000 & 0.000 \\
\hline
\multirow{3}{*}{1.2B}
& 300B & 500B & 0.758 & 0.229 & 0.012 & 0.001 & 0.000 & 0.000 \\
& 500B & 500B & 0.619 & 0.376 & 0.004 & 0.001 & 0.000 & 0.000 \\
& 800B & 500B & 0.496 & 0.492 & 0.007 & 0.003 & 0.002 & 0.000 \\
& 1000B & 500B & 0.496 & 0.492 & 0.007 & 0.003 & 0.002 & 0.000 \\
\hline
\end{tabular}
\end{table*}

\section{FITTING EXPERIMENTS}

\subsection{Training setup}
\label{sec:Training-setup}

We train 9 models ranging from 252M to 1.2B on 3 layermix sampling weights $HQ$, $MQ$, and $LQ$, with 3.6x over-trained ratio, resulting in 27 experiment runs in total to collect data for fitting the InfoLaw parameters. We use transformer architecture \citep{Attention}, SwiGLU \citep{SwiGLU} as the activation function and RoPE embeddings \citep{ROPE}. We use a tokenizer with 250k vocabulary. See Appendix \ref{app:layermix} and Appendix \ref{app:training} for details about LayerMix sampling weights, model structure, learning rate and optimizer.

\subsection{Fitting the curve}
\label{sec:fit-curve}

In this section, we introduce how to fit the parameters in InfoLaw to predict the model performance collected in Section \ref{sec:Training-setup}. Since Information $info$ indicates the knowledge learned by the model, we expect larger $info$ to correspond to lower evaluation loss $L$. Considering that there may exist scale difference between $info$ and model loss $L$, we choose Spearman correlation $\rho_s$ as the fitting metric, i.e., the object is to find the optimal quality density $f$ and $\lambda(N)$ such that the Spearman correlation between evaluation loss $L$ and $info$ is minimized for all the experiments over $N,w$ :

\begin{equation}
\label{eq:argmax_f}
\begin{aligned}
(f^*, \lambda^*) = \underset{f, \lambda}{\operatorname{argmin}} \sum_{N, w} \rho_s \big(L_N,
\text{info}(w, K_N, S_N, f, \lambda(N)) \big)
\end{aligned}
\end{equation}

To prevent from over-fitting, we make some assumption based on naive intuition. For $f$, as it indicates the quality density, the higher-quality bucket should have larger $f$. As smaller $d$ corresponds to higher-quality buckets, we define $f$ in the following form to ensure it is a decreasing function:

\begin{equation}
f_d(\theta) = e^{-\theta * d}
\label{eq:f}
\end{equation}

where $\theta$ is a hyperparameter and $\theta>0$. 

$\lambda(N)$ is related to the model's learning capacity, so $\lambda(N)$ should increase as $N$ increases. But we need to find the formula for $\lambda(N)$ related with $N$ so that it can scale to larger $N$. To do this we first sample 100,000 combinations of $\theta$ and $\lambda(N)$ from the parameter space, then select optimal $\theta^*$ and $\lambda^*_N$ based on Equation \ref{eq:argmax_f}. The fitted quality density $f(\theta^*)$ is shown in Figure \ref{fig:quality_density} with fitted $\theta^*=0.922$.

Having the $\lambda(N)^*$ values of different models, as is shown in Figure \ref{fig:lmbd-N}, we try to fit the $\lambda(N)$-$N$ curve. The relationship between $\lambda(N)$ and $N$ is observed to be non-linear, exhibiting rapid growth for smaller $N$ and gradually saturating as $N$ increases. This trend is well-approximated by a logarithmic function. Therefore, we choose the $\lambda(N)$-$N$ curve using following formula:

\begin{equation}
\lambda(N)(a,b) = a \cdot \ln(N) + b
\label{eq:logarithmic-func}
\end{equation}

Using existing $\lambda(N)^*$, we fit the $\lambda(N)$-$N$ curve in Figure \ref{fig:lmbd-N} with fitted $a^*=0.140$, $b^*=0.018$. To validate this fit, we compute $\lambda(N)^*$ for larger $N$ under the fixed $\theta^*$, and examine whether these values lie on the predicted $\lambda(N)$-$N$ curve. As illustrated in Figure~\ref{fig:lmbd-N}, the results demonstrate strong extrapolation performance, supporting the correctness of our formulation. We compared with different formats of \ref{eq:logarithmic-func} in Appendix \ref{app:lambda} and the log function best fit the trend and extrapolates.

Finally, with $f(\theta^*)$ and $\lambda(N)(a^*,b^*)$, we can calculate the Information for arbitrary layermix sampling weights $w$, train token $K$, source token $S$ and model non-embedding FLOPs/token $N$.

\begin{figure*}
    \centering
    \includegraphics[width=0.85\textwidth]{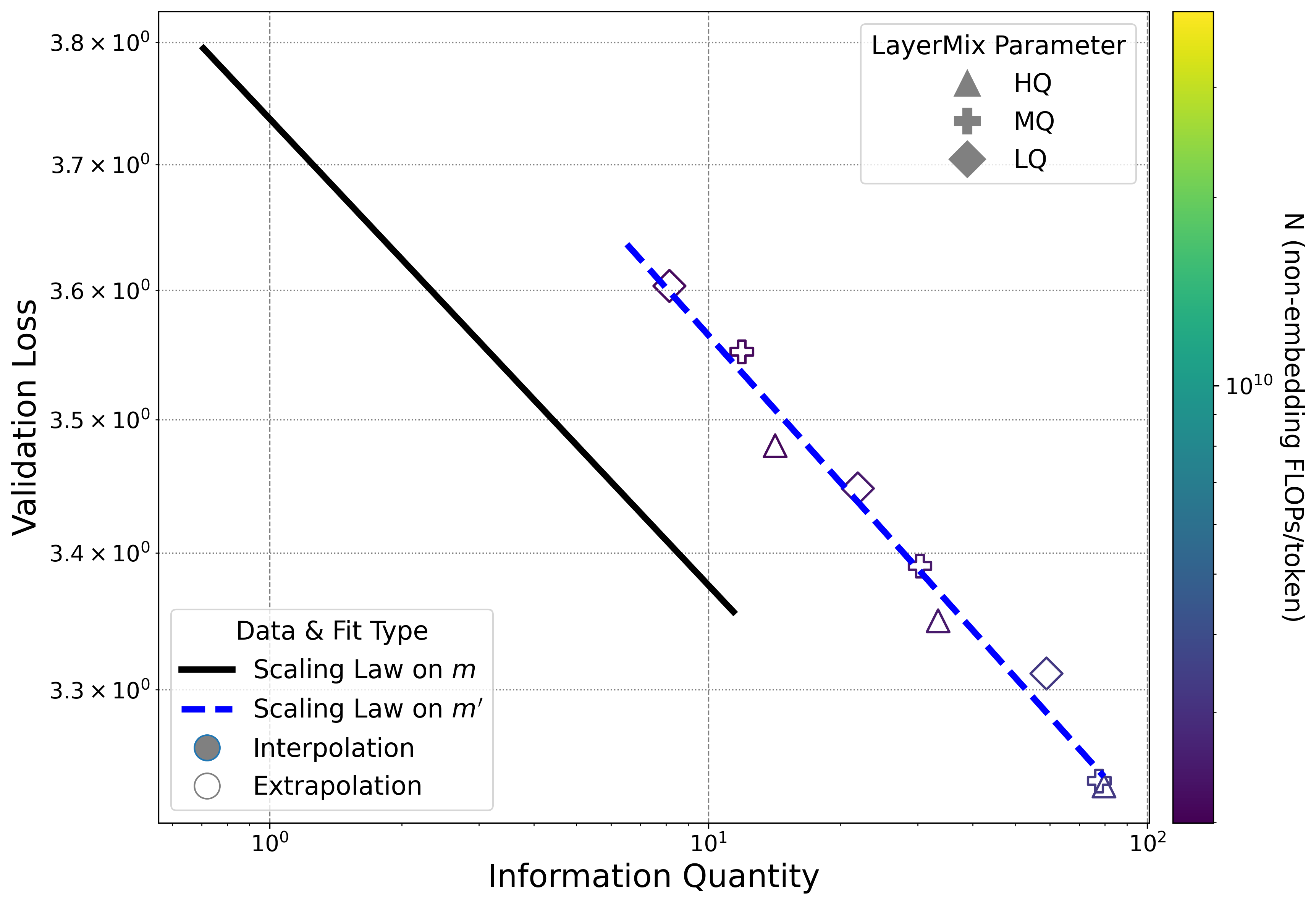}
    \caption{Cross-Regime Prediction of the Scaling Law. The blue line ($C_{m'}$) is a pure prediction, generated using parameters fitted only on the $C_m$ data (black line). The fit for the $C_{m'}$ points demonstrates our InfoLaw's power to extrapolate across different overtrain degrees.}
    \label{fig:extra_600B}
\end{figure*}

\section{Extrapolation}


After fitting InfoLaw on the 252M--1.2B models, we evaluate its extrapolation along three axes: unseen mixture recipes, larger model scales, and a higher overtraining ratio. Finally, we use our InfoLaw to predict optimal data recipe under different training budgets and validate the optimal recipe by comparing with preset recipes.

\noindent{\textbf{Comparing with traditional scaling laws}}

Figure \ref{fig:info_vs_scalinglaw} contrasts our InfoLaw with traditional power scaling law in the loss–$C$ plane. Both curves are fit using models in the 252M–1.2B range and then extrapolated to larger models. The Info curve tracks the MLQ data more closely within the fitting regime and remains accurate when extrapolating up to 7B models, avoiding the overly optimistic loss reductions predicted by the traditional law at high compute. Concretely, the traditional scaling law tends to under-estimate loss as $C_m$ grows, whereas the Info curve better matches the realized validation losses of larger models.

\noindent{\textbf{Extrapolation to other LayerMix Sampling Weights}}

We first test the ability to generalize to an unseen LayerMix sampling weights. We test on unseen dataset generated with $MLQ$, $MHQ$ on model scales ranging from $252M$ to $1.2B$, which are within the range of training data. Also we random sample 25 more sampling weights and run experiments on $1.2B$ model only. 

The result is shown in Figure \ref{fig:scaling_laws_all} . As can be seen, these points align remarkably well with the scaling law curve established by the initial $HQ$, $MQ$, $LQ$ data, demonstrating the predictive power of our model on unseen LayerMix sampling weights. The traditional scaling laws requires additional experiments on different data recipes to fit new curves, while ours can directly predict loss on unseen recipes. 

\noindent{\textbf{Extrapolation to Larger Models}}

To test the extrapolation ability on model scale, we use the same Layermix sampling weights $MQ$,$LQ$ to train models ranging from 1.5B to 2.5B and $HQ$, $LQ$ to train model with 2.5B parameters, which are out of the range of training data. The experimental results of larger models are shown in Figure \ref{fig:scaling_laws_all}(a-e), we can see InfoLaw predict the loss on larger scale accurately for all three sampling weights, proving the ability of scaling on model size.

\noindent{\textbf{Combination of Extrapolation}}

Furthermore, we combine the two extrapolation above and test the effectiveness on both unseen LayerMix sampling weights and unseen scales. We run experiments with $MLQ$,$MHQ$ on models ranging from $1.5B$ to $7B$. As shown in Figure \ref{fig:infolaw}, InfoLaw also generalise well on these combined extrapolation condition. On all the unseen data points, including unseen LayerMix sampling weights ($MLQ$, $MHQ$ and other 25 sets random sampled weights) and unseen model scales, InfoLaw predict the validation loss with $0.15\%$ average absolute error and maximum error is $0.96\%$. This proves that our proposed information scaling law has reliable extrapolation capability. 

\noindent{\textbf{Extrapolation to Larger Overtrain Degree}}

To explore the model's reliability under varying sub-optimality, we conducted a second series of experiments at a higher overtrain degree, $m'=25$. This new regime was anchored by a 1.2B model trained on 640B tokens (the $C_{m'}$ experiment), contrasting with our initial $C_m$ experiment anchored at 106B tokens.

For the $C_{m'}$-experiment, we calculated the Information using the same quality density $f(\theta^*)$ and $\lambda(N)(a^*,b^*)$ fitted previously on the $C_m$ data. As shown in Figure \ref{fig:extra_600B}, the new experimental points align with a new scaling law curve. The resulting curves for $C_m$ and $C_{m'}$ appear nearly parallel, suggesting the overtrain degree $m$ primarily shifts the curve's intercept. This confirms that our proposed Information Scaling Law is effective across different overtrain degrees.

\noindent{\textbf{Optimizing Data Recipe with InfoLaw}}

The ability of predicting loss on unseen data recipes and scales enables us to search for best data recipe without additional experiments. Similar to \citet{RegMix}. We randomly sample 100k LayerMix parameters from the parameter space, compute the information for each set of parameters, and convert it to loss via Equation \ref{eq:loss-C}. We then select the parameter that minimizes the predicted validation loss as the optimal LayerMix configuration for each training setting.
To verify the optimal recipe, we conduct experiments on $2.5B$ model with optimal data recipe and 3 other Layermix sampling weights. The result optimal recipe is as in Table \ref{tab:layermix_param}. As shown in Figure \ref{fig:result_2B5}, our optimal recipe achieves the best validation loss. 

We additionally test generalization to unseen LayerMix parameters: on 25 held-out configurations for the 1.2B model, predicted and measured validation losses achieve a Pearson correlation of $0.76$, suggesting InfoLaw can reliably rank recipes for efficient search.

In Table \ref{tab:best_mix}, we present the optimal LayerMix parameters for different model sizes and training-token counts under a fixed source-token budget of 500B tokens. The optimal LayerMix parameters exhibit two clear trends. First, at a fixed training-token count, smaller models favor a higher fraction of high-quality data, whereas larger models benefit more from diversity and thus allocate a smaller fraction to the high-quality data. Second, as the total training tokens increase, the optimal LayerMix parameters shift from a high-quality emphasis toward greater diversity. More results are shown in Appendix \ref{app:layermix-pred}. In short: Small models or small training budgets prioritize quality; large models or large training budgets prioritize diversity.

\section{Conclusion}

In this paper, we propose a refined scaling law modeling \textbf{InfoLaw}, which focus on predicting model performance on downstream tasks under data-constrained settings with weighted-quality mixing. The InfoLaw provides accurate predictions of model performance on unseen data recipes at larger computational scales, achieving an average absolute error of only 0.15\% and a maximum error of 0.96\%. This enables efficient discovery of optimal data recipes without the need for extensive additional experiments. Furthermore, the InfoLaw extrapolates reliably across varying degrees of over-training, offering an effective tool for selecting data recipes under different computational budgets. 

\section{Impact Statement}
This paper aims to advance machine learning by improving our understanding of large language model performance under different data mixing and repetition strategies. Our InfoLaw can support more efficient pretraining by reducing expensive trial-and-error over data recipes. We do not anticipate direct negative societal consequences arising uniquely from this contribution. Broader ethical issues associated with LLMs, such as bias, misuse, and unsafe deployment, remain important but are not specifically introduced or materially amplified by our method beyond general improvements in training efficiency.

\bibliography{example_paper}
\bibliographystyle{icml2026}

\newpage
\appendix
\onecolumn

\section{Training Dataset}
\label{app:train-set}

We use the English portion of the Common Crawl Dataset \citep{commoncrawl}, utilizing 96 of the snapshots, from CC-MAIN-2013-20 to CC-MAIN-2024-18. Following \citet{deepseek}, we ran a global fuzzy deduplication across all snapshots, resulting in a total dataset with 3.7T tokens.

\section{Justification for the Normalization Term \texorpdfstring{$\log(K)$}{log(K)}}
\label{app:normalization_justification}

In Equation~\ref{eq:K_part} , we incorporate a normalization term $\log(K)$ into the decay function to model the interaction between repetition decay and the total token budget. We selected this logarithmic form after rigorously evaluating alternative formulations. Specifically, we compared our chosen decay term against constant normalization and power-law normalization:

\begin{itemize}
    \item \textbf{Constant Normalization:} Assuming the decay rate is independent of the dataset scale:
    \begin{equation}
        \text{Decay}(t) \propto e^{-\lambda(N) t}
    \end{equation}
    \item \textbf{Power-Law Normalization:} Assuming the decay scales polynomially with the token budget:
    \begin{equation}
        \text{Decay}(t) \propto e^{-\frac{\lambda(N) t}{K^\alpha}}
    \end{equation}
    \item \textbf{Logarithmic Normalization (Ours):}
    \begin{equation}
        \text{Decay}(t) \propto e^{-\frac{\lambda(N) t}{\log(K)}}
    \end{equation}
\end{itemize}

While we omit the visual plots for brevity, our preliminary experiments demonstrated that the alternative forms failed to unify the scaling behaviors across different token budgets $K$:

\begin{enumerate}
    \item \textbf{Failure of Constant Normalization:} This formulation fails to account for the scaling properties of information density. Empirically, we observed that it systematically {overestimates the accumulated Information} for large models when trained with larger token budgets. Consequently, this leads to overly optimistic loss predictions that deviate significantly from the actual experimental results.
    
    \item \textbf{Failure of Power-Law Normalization:} We found this formulation to be fundamentally unsuitable. It resulted in a {complete failure to fit the relationship between Information and Validation Loss}. The data points derived using power-law normalization remained scattered without exhibiting the necessary power-law correlation, rendering it impossible to derive a valid scaling law.
\end{enumerate}

In contrast, the $\log(K)$ term was the only formulation that minimized the alignment error—successfully collapsing diverse configurations of $(w, K, S)$ onto a single unified power-law curve (as shown in Figure~\ref{fig:infolaw})—and maintained a low extrapolation error across the full range of model scales (252M to 7B). This suggests that the marginal utility of repeated data diminishes logarithmically relative to the total training budget.

\section{LayerMix Sampling Function}
\label{app:layermix}

We show the detail of LayerMix sampling function in Algorithm \ref{alg:layermix}.

\begin{algorithm}[h]
\caption{LayerMix Sampling Function $H(w, K, S, B)$}
\label{alg:layermix}
\DontPrintSemicolon
\SetKwProg{Fn}{Function}{:}{end}
\SetKwInOut{Input}{Input}
\SetKwInOut{Output}{Output}
\SetKwFunction{FH}{H}

\Fn{\FH{$w, K, S, B$}}{
    \Input{
        $w$: list of target proportions for six buckets, $w=[w_0,\dots,w_5]$, $\sum w_d = 1$\;
        $K$: total number of tokens for the final training dataset\;
        $S$: total number of tokens in the entire source corpora\;
        $B$: source distribution proportions $B=[0.05, 0.15, 0.2, 0.2, 0.2, 0.2]$\;
    }
    \Output{
        $D_{train}$: final packed training dataset\;
        $M$: list of unique token counts per layer, $M=[M_0,\dots,M_5]$\;
        $R$: list of average repetition counts per layer, $R=[R_0,\dots,R_5]$\;
    }
    \BlankLine
    Initialize empty training dataset $D_{train} \gets \emptyset$\;
    Initialize empty statistics lists $M \gets [],\ R \gets []$\;
    \BlankLine
    \For{$d \gets 0$ \KwTo $5$}{
        \tcp{Iterate through each quality bucket}
        $K_{needed} \gets K \times w_d$ \tcp*{tokens needed from bucket $d$ for the target mix}
        $S_d \gets S \times B[d]$ \tcp*{source tokens available in bucket $d$}
        $Ratio_d \gets K_{needed} / S_d$ \tcp*{sampling ratio for current bucket}
        \BlankLine
        \tcp{Detailed sampling process for bucket $d$}
        Initialize empty temporary set $D_{sampled\_d} \gets \emptyset$\;
        \ForEach{data point $x$ in bucket $d$}{
            \tcp{1. Deterministic copy for the integer part of the ratio}
            \For{$i \gets 1$ \KwTo $\lfloor Ratio_d \rfloor$}{
                Add $x$ to $D_{sampled\_d}$\;
            }
            \tcp{2. Probabilistic sampling for the fractional part}
            \If{$Ratio_d - \lfloor Ratio_d \rfloor > 0$ \textbf{and} $random() < (Ratio_d - \lfloor Ratio_d \rfloor)$}{
                Add $x$ to $D_{sampled\_d}$\;
            }
        }
        Append all data from $D_{sampled\_d}$ to $D_{train}$\;
        \BlankLine
        $M_d \gets \min(K_{needed}, S_d)$ \tcp*{unique tokens for bucket $d$}
        Append $M_d$ to $M$\;
        $R_d \gets K_{needed} / M_d$ \tcp*{average repetition count}
        Append $R_d$ to $R$\;
    }
    \BlankLine
    \Return $D_{train}, M, R$ \tcp*{dataset and statistics}
}
\end{algorithm}

\begin{equation}
\sqrt m = \frac{N_{opt}}{N}=\frac{D}{D_{opt}}
\label{eq:overtrain_m}
\end{equation}

A value of $m=1$ indicates a compute-optimal training run, while $m>1$ signifies that the model is overtrained relative to its compute budget.

\begin{algorithm}[h]
\setcounter{AlgoLine}{0}   
\caption{Calculation of Overtrain Degree and Optimal Tokens}
\label{alg:overtrain_calc}
\DontPrintSemicolon
\SetKwProg{Fn}{Function}{:}{end}
\SetKwInOut{Input}{Input}
\SetKwInOut{Output}{Output}
\SetKwFunction{FCalc}{CalculateOvertrainExtrapolation}
\SetKwFunction{FGetN}{Get\_N}

\Fn{\FCalc{$model_{curr}, D_{curr}, models_{target}$}}{
    \Input{
        $model_{curr}$: size of the current model configuration\;
        $D_{curr}$: number of tokens used to train the current model\;
        $model_{target}$: size of the target model configuration\;
    }
    \Output{
        $m$: calculated overtrain degree for the current configuration\;
        $D_{target}$: train tokens of target model under the same overtrain degree\;
    }
    \BlankLine
    \tcp{Part 1: Calculate overtrain degree $m$ from the current configuration}
    $N_{curr} \gets \FGetN(model_{curr})$ \tcp*{non-embedding FLOPs/token for current model}
    $C \gets N_{curr} \times D_{curr}$ \tcp*{total compute budget}
    \BlankLine
    $N_{opt} \gets 0.06085 \times C^{0.5445}$ \tcp*{Chinchilla-optimal $N$ for budget $C$}
    $D_{opt} \gets 16.4326 \times C^{0.4555}$ \tcp*{Chinchilla-optimal tokens for budget $C$}
    \BlankLine
    $\sqrt{m} \gets N_{opt} / N_{curr}$ \tcp*{overtrain degree $m$ (equiv. $\sqrt{m}=D_{curr}/D_{opt}$)}
    \BlankLine
    \tcp{Part 2: Extrapolate to target model while keeping $m$ constant}
    \ForEach{$model_{t}$ \KwTo $[model_{curr}] + models_{target}$}{
        $N_{t} \gets \FGetN(model_{t})$ \tcp*{non-embedding FLOPs/token for target model}
        $N'_{opt} \gets N_{t} \times \sqrt{m}$ \tcp*{corresponding optimal $N$ for the target}
        $C_{new} \gets (N'_{opt} / 0.06085)^{1/0.5445}$ \tcp*{derive new compute budget}
        $D'_{opt} \gets 16.4326 \times C_{new}^{0.4555}$ \tcp*{optimal tokens for the new budget}
        $D_{target} \gets D'_{opt} \times \sqrt{m}$ \tcp*{required tokens for the target model}
    }
    \BlankLine
    \Return $m, D_{target}$ \tcp*{overtrain degree and target train tokens at same $m$}
}
\end{algorithm}

\section{Training}
\label{app:training}

The model structures used in LayerMix are illustrated in Table \ref{tab:model-structure}. We train all the model with 2048 as the max sequence length, we use a cosine decay scheduler and the initial learning rate calculated by $lr=round(0.3118\cdot C^{-0.1250}, 8)$, the warm up ratio is set 0.5\%. We use AdamW optimizer with $\beta_1=0.9$, $\beta_2=0.95$, weight decay$=0.1$. 

\begin{table}[h]
    \caption{Structure of models used in LayerMix.}
    \label{tab:model-structure} 
    \centering
    \begin{tabular}{l|cccc}
        \hline
        \textbf{Model} & \textbf{Hidden dim. (C)} & \textbf{MLP dim. (D)} & \textbf{Layers (L)} & \textbf{Heads} \\  
        \hline
        \textbf{252M}  & 1024 & 2752  & 20 & 16 \\
        \textbf{302M}  & 1024 & 2752  & 24 & 16 \\
        \textbf{392M}  & 1280 & 3392  & 20 & 20 \\
        \textbf{470M}  & 1280 & 3392  & 24 & 20 \\
        \textbf{566M}  & 1536 & 4096  & 20 & 24 \\
        \textbf{680M}  & 1536 & 4096  & 24 & 24 \\
        \textbf{850M}  & 1792 & 4800  & 22 & 28 \\
        \textbf{1B}    & 1920 & 5120  & 24 & 30 \\
        \textbf{1.2B}  & 2048 & 5440  & 24 & 16 \\
        \textbf{1.5B}  & 2304 & 6144  & 24 & 36 \\
        \textbf{1.8B}  & 2304 & 6144  & 28 & 36 \\
        \textbf{2.5B}  & 2560 & 6848  & 32 & 40 \\
        \textbf{7.7B}  & 4096 & 14336 & 32 & 32 \\
        \hline
    \end{tabular}
\end{table}

\section{Supplementary Analysis of Repetition Effects}
\label{app:motivation_figs}

\begin{figure*}
    \centering
    \begin{subfigure}[t]{0.45\textwidth}
        \centering
        \includegraphics[width=\textwidth]{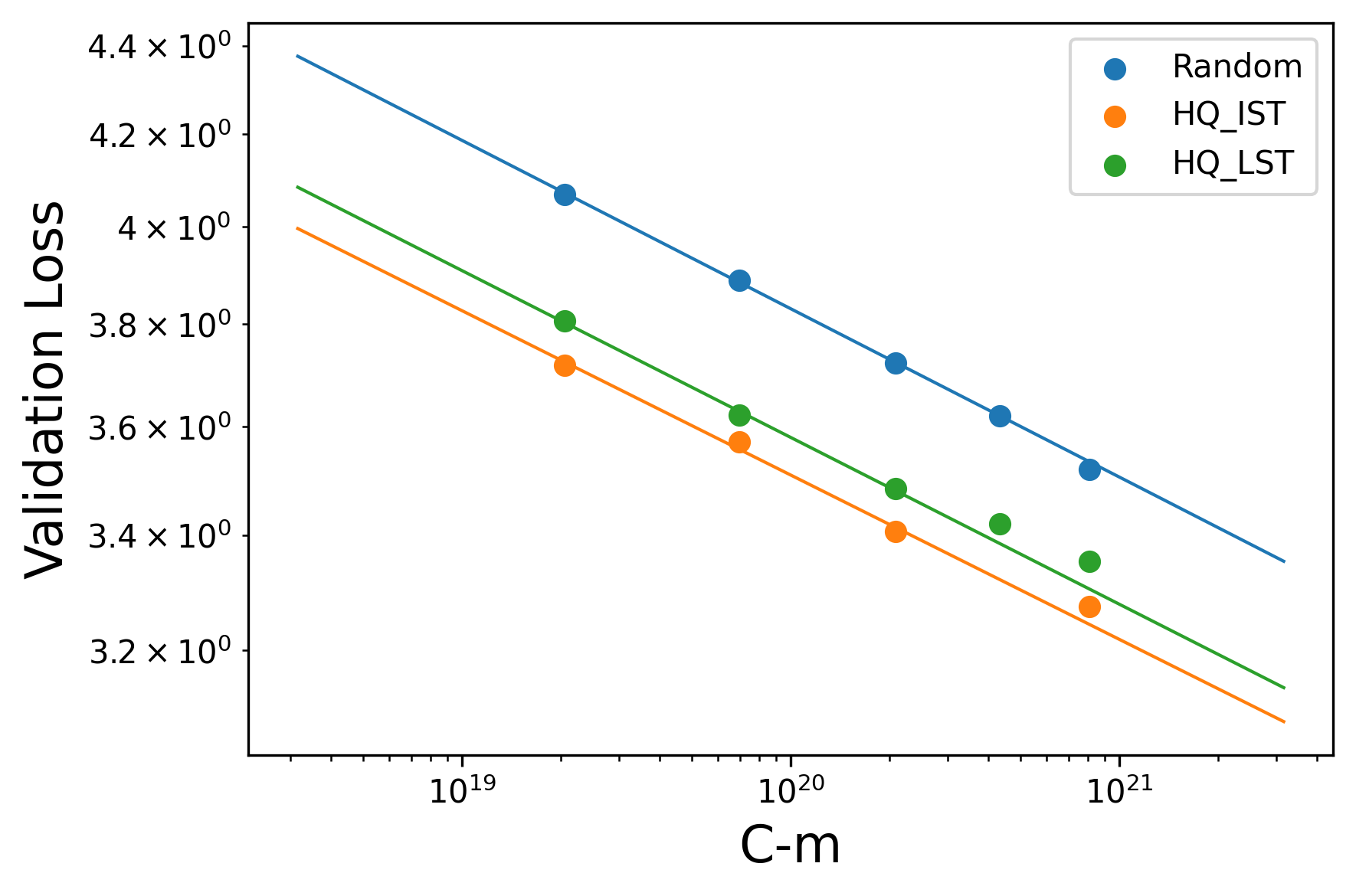}
        \caption{Loss--$C_m$ curves under different data regimes. Random: large source ($S\!\gg\!K$) with negligible repetition. HQ\_IST: LayerMix with the HQ recipe and $S\!\gg\!K$ (negligible repetition). HQ\_LST: the same HQ recipe but $S\!=\!K$, inducing repetition.}
        \label{fig:app_random_ksn_v1v2}
    \end{subfigure}
    \hfill
    \begin{subfigure}[t]{0.45\textwidth}
        \centering
        \includegraphics[width=\textwidth]{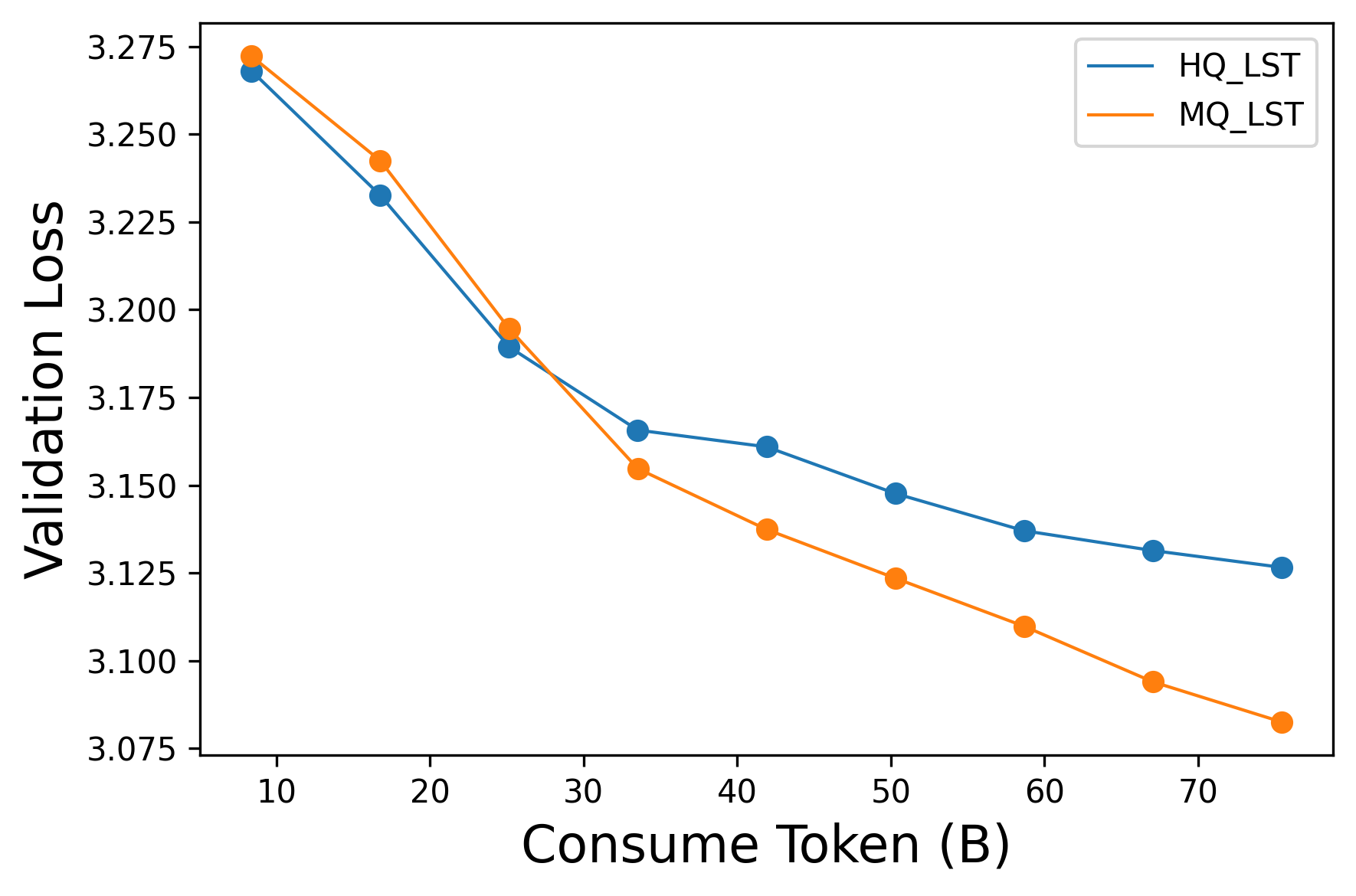}
        \caption{Training-time evaluation loss for two 850M runs (HQ\_LST vs MQ\_LST), illustrating late-stage slowdown under heavier repetition.}
        \label{fig:app_cross_v2}
    \end{subfigure}
    \caption{\textbf{Supplementary evidence for repetition effects.} (a) In the loss--$C_m$ view, repetition induces systematic deviation from a single power-law trend. (b) Heavier repetition leads to slower late-stage improvement and worse final loss, consistent with diminishing returns.}
    \label{fig:app_motivation}
\end{figure*}

\paragraph{Notation.}
IST (Infinite Source Tokens) denotes $S\!\gg\!K$, where repetition is negligible; LST (Limited Source Tokens) denotes $S\!=\!K$, where repetition is induced by the sampling weights. HQ/MQ refer to the LayerMix preset recipes in Table~\ref{tab:layermix_param}. Figure~\ref{fig:app_motivation}(a) provides an additional sanity check for the loss--$C_m$ behavior under different repetition regimes, while Figure~\ref{fig:app_motivation}(b) shows the corresponding training-time dynamics motivating a saturation/decay model.

\section{The relationship between benchmark validation loss and performance}
\label{app:val-loss}

\begin{figure}
    \centering
    \includegraphics[width=\textwidth]{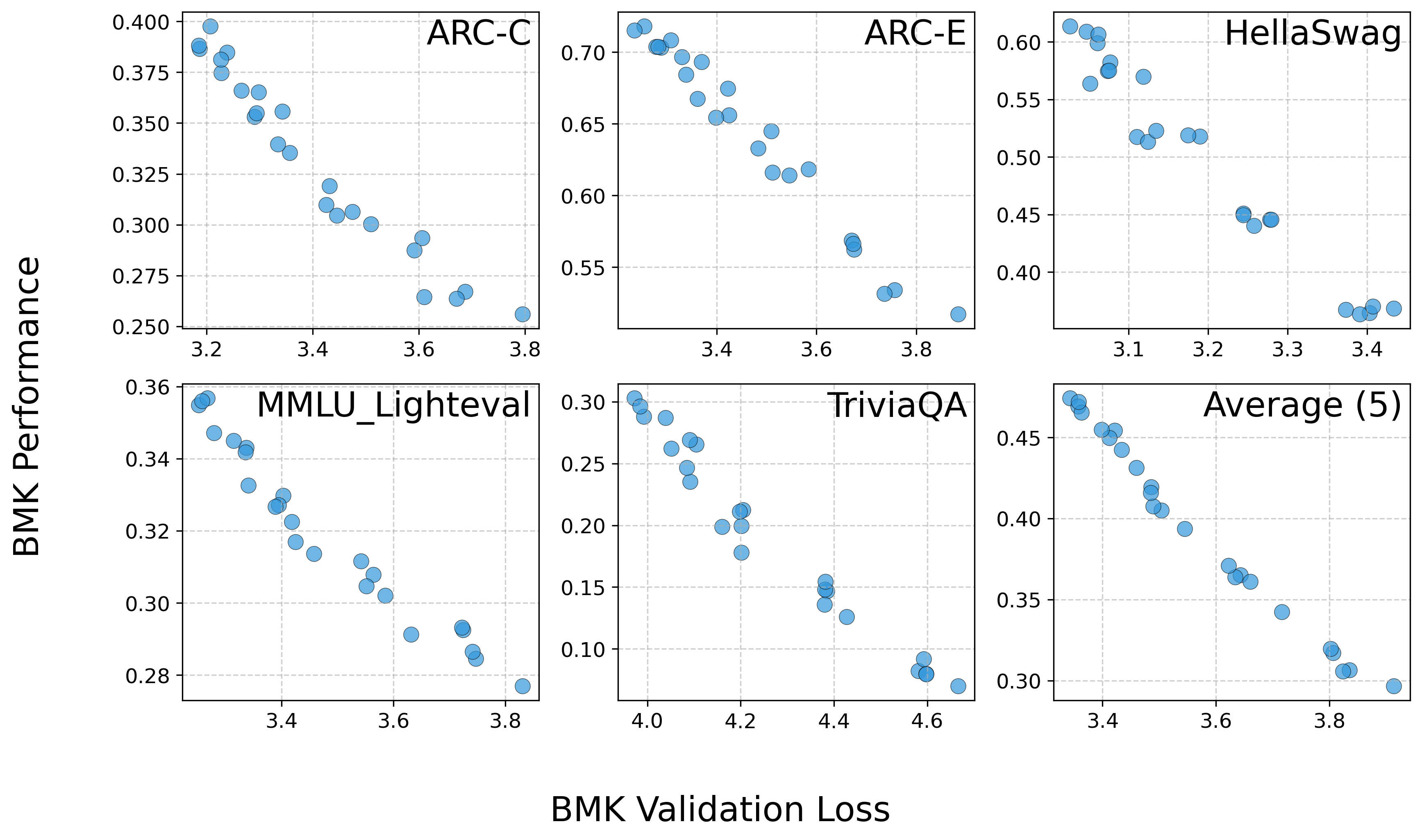}
    \caption{Validation loss versus downstream performance across benchmarks (ARC-C, ARC-E, HellaSwag, MMLU-Lighteval, TriviaQA) and their average.}
    \label{fig:loss_perf}
\end{figure}

Our InfoLaw focus on predicting the evaluation loss on downstream benchmarks. However, it also represents for the actual downstream performance. Figure \ref{fig:loss_perf} shows a near-linear relationship between validation loss and downstream performance on our evaluation tasks, and Table \ref{tab:spearman_loss_perf} shows the spearman corelation between validation loss and downstream performance. Lower loss consistently corresponds to higher performance within the operating regime of our models. This indicates that improvements in loss provide reliable signals for expected gains in downstream performance.

\begin{table}[t]
    \centering
    \caption{Spearman correlation between validation loss and performance across benchmarks}
    \label{tab:spearman_loss_perf}
    \begin{tabular}{lcc}
        \hline
        Benchmark & Spearman $r_s$ & $p$-value \\
        \hline
        ARC-C        & -0.979 & $1.02 \times 10^{-16}$ \\
        ARC-E        & -0.982 & $2.72 \times 10^{-17}$ \\
        HellaSwag    & -0.942 & $6.13 \times 10^{-12}$ \\
        MMLU-LightEval        & -0.989 & $1.26 \times 10^{-19}$ \\
        TriviaQA     & -0.970 & $4.53 \times 10^{-15}$ \\
        Average (5)  & -0.996 & $3.54 \times 10^{-24}$ \\
        \hline
    \end{tabular}
\end{table}

\section{Alternative Fits for $\lambda$}
\label{app:lambda}

\begin{figure}
    \centering
    \includegraphics[width=0.7\textwidth]{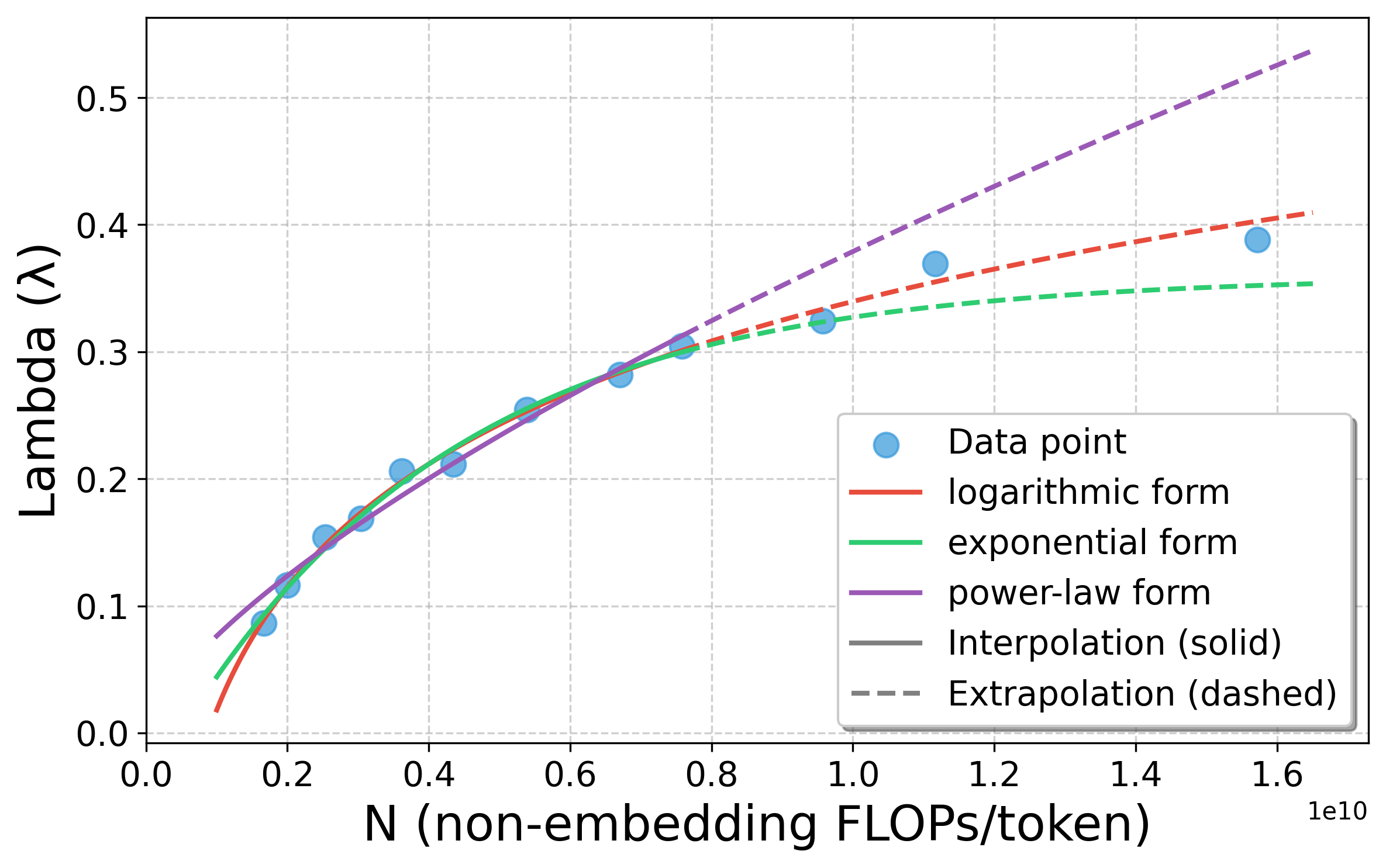}
    \caption{Comparison of functional fits for $\lambda$ as a function of $N$ (non-embedding FLOPs/token). The logarithmic form provides the best in-domain fit and extrapolation behavior compared with the exponential and power-law alternatives. Solid lines denote interpolation over observed $N$; dashed lines indicate extrapolation beyond the observed range.}
    \label{fig:lambda_fit_3_curve}
\end{figure}

In Section \ref{sec:fit-curve}, we model the relationship between non-embedding FLOPs/token $N$ and hyperparameter $\lambda$. Our primary specification adopts the logarithmic form Equation \ref{eq:logarithmic-func}. Beyond this baseline, we also evaluated alternative function families, including an exponential form:
\begin{equation}
\lambda(x; a, b, c) = a \cdot \bigl(1 - e^{-b x + c}\bigr)
\label{eq:exp-func}
\end{equation}
and a power-law form:
\begin{equation}
\lambda(x; a, b) = a \cdot x^{b}
\label{eq:power-func}
\end{equation}
As shown in Figure \ref{fig:lambda_fit_3_curve}, the logarithmic model achieves the best fit to the $N-\lambda$ relationship, outperforming the exponential and power-law alternatives. Accordingly, we adopt function \ref{eq:logarithmic-func} as the final parameterization.

\section{Deviation of Traditional Scaling Law}
\label{app:TraditionalSL}

\begin{figure}
    \centering
    \includegraphics[width=\textwidth]{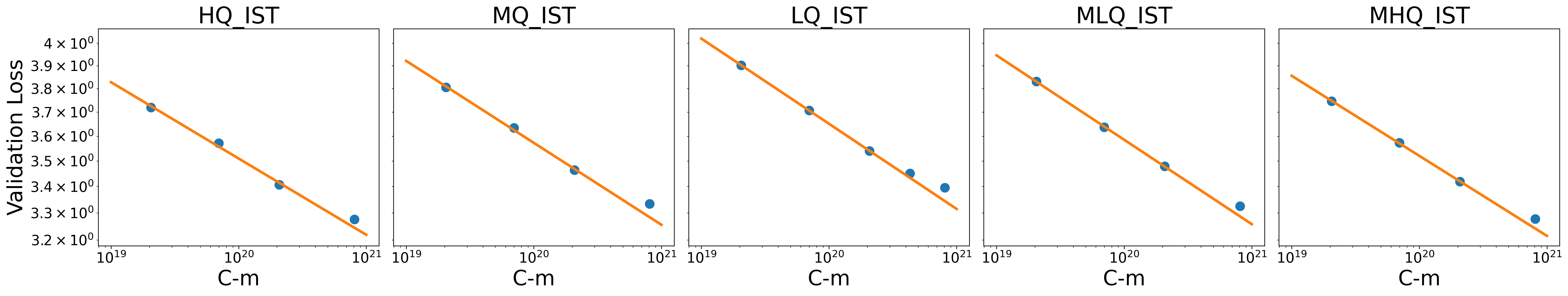}
    \caption{$Loss$ and $C_m$ Curve of different LayerMix $IST$ experiments}
    \label{fig:scalinglaw_v1_all}
\end{figure}

\begin{figure}
    \centering
    \includegraphics[width=\textwidth]{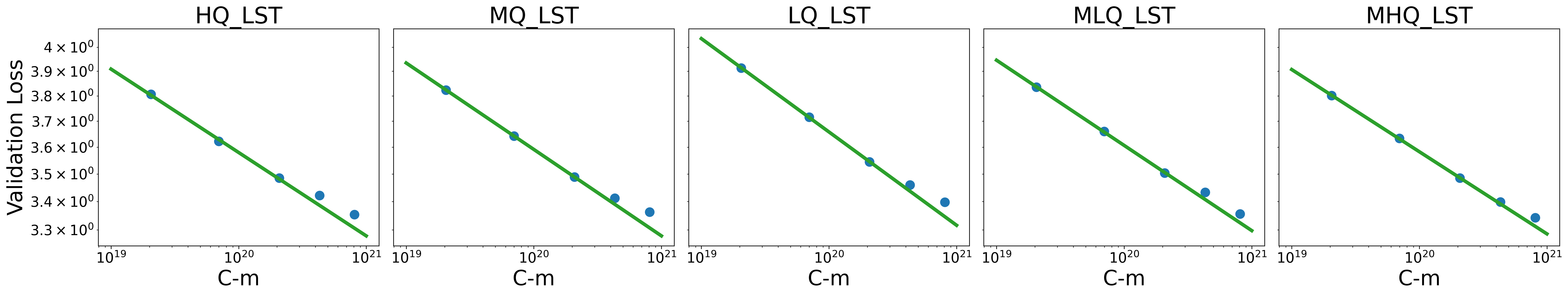}
    \caption{$Loss$ and $C_m$ Curve of different LayerMix $LST$ experiments}
    \label{fig:scalinglaw_v2_all}
\end{figure}

We show all $Loss$-$C$ curve of different LayerMix sampling weights with IST and LST in Figure \ref{fig:scalinglaw_v1_all} and Figure \ref{fig:scalinglaw_v2_all}, they all exhibit a clear deviation from the traditional scaling law, which is fitted from the first three data points.

\section{Quality Score}

\begin{figure}
    \centering
    \includegraphics[width=\textwidth]{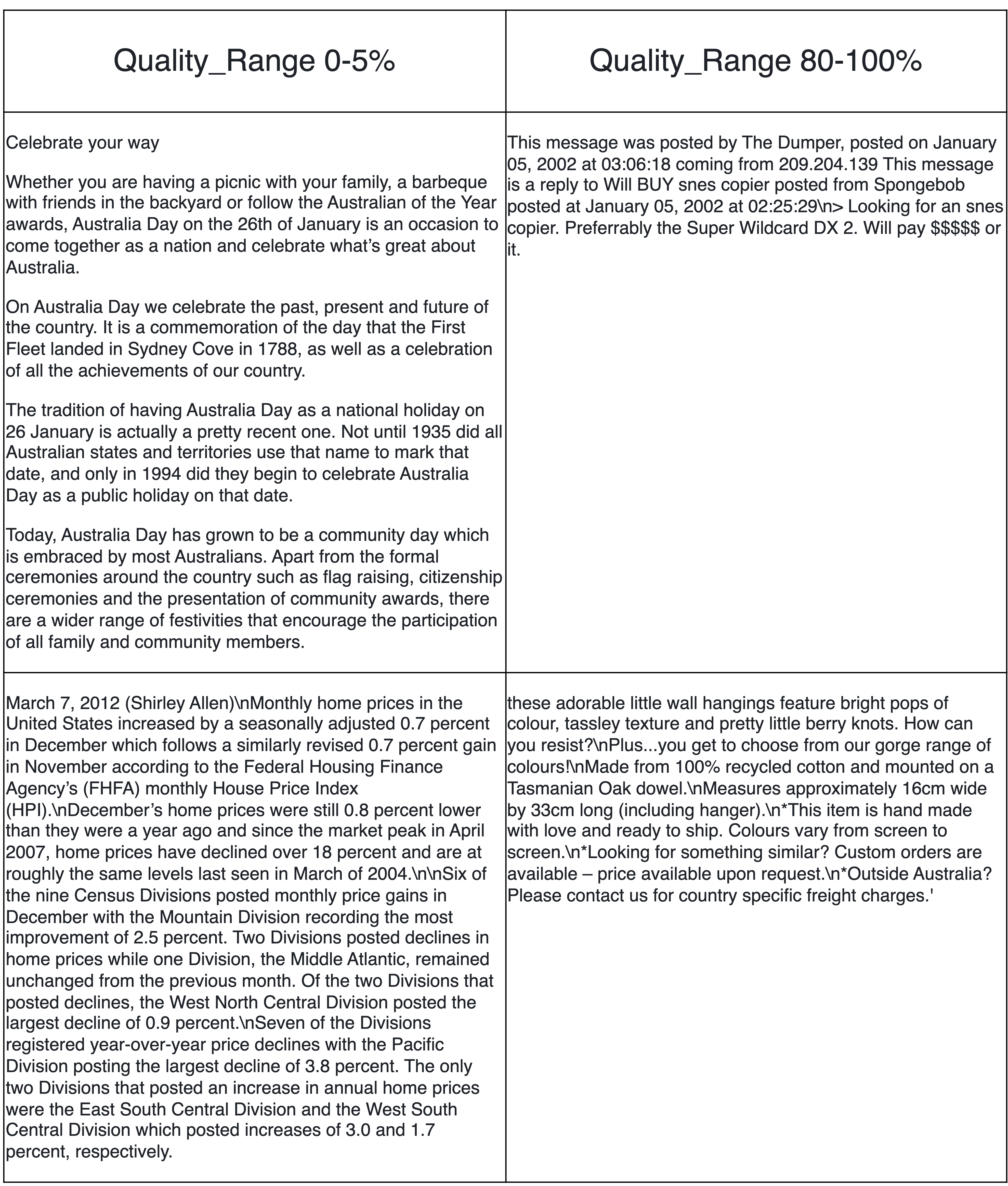}
    \caption{Case study contrasting data quality. Left (0–5\% quality range): coherent, informational, and instructional passages. Right (80–100\% quality range): low-information, ad-like content with minimal reasoning or educational value.}
    \label{fig:case_study}
\end{figure}

We show some data samples in different Quality buckets in Figure \ref{fig:case_study}. This figure indicates that high-score samples under our merged FineWebEdu and DCLM scores are more coherent and instructional. By contrast, low-score cases predominantly consist of advertisements or low-information content, offering little substantive value.

Table \ref{tab:finewebedu_exp} reports four benchmark results for training a 1.2B model from scratch on 30B tokens using three datasets: the top 5\% and top 20\% selected by the FineWebEdu classifier, and a random sample, all from \citet{RefinedWeb}. High-quality data selected by FineWebEdu outperforms the random baseline, and higher-quality subsets yield better results.

\begin{table}
\centering
\caption{FineWebEdu-selected subsets vs. random data for training a 1.2B model on 30B tokens}
\label{tab:finewebedu_exp}
\begin{tabular}{cccccc|c}
\hline
Model & Data & ARC-C & HellaSwag& TriviaQA& MMLU-LightEval& avg \\
\hline
1.2B & Random 30B      &28.50\% & 51.56\% & 15.55\% & 30.23\% & 31.46\% \\
\hline
1.2B & FWE-top20\% 30B &34.30\% & 55.26\% & 20.05\% & 32.82\% & 35.61\% \\
1.2B & FWE-top5\%  30B &37.20\% & 55.14\% & 19.25\% & 34.50\% & 36.52\% \\
\hline
\end{tabular}
\end{table}

\section{Optimizing Token Mix with InfoLaw}
\label{app:layermix-pred}

We present the detailed optimal LayerMix parameters (or token-mix ratios) for different models and training budgets predicted by InfoLaw in Table \ref{tab:model_data}. This table shows that small models or small training budgets prioritize quality, while large models or large training budgets prioritize diversity.

\begin{table}
\centering
\caption{The detailed best layer token mix for different models and train token}
\label{tab:model_data}
\begin{tabular}{cccccccccc}
\hline
\textbf{Model} & \textbf{Train Token} & \textbf{Source Token} & \textbf{$w0$} & \textbf{$w1$} & \textbf{$w2$} & \textbf{$w3$} & \textbf{$w4$} & \textbf{$w5$} \\
\hline
\multirow{9}{*}{7B} & 200B & 500B & 0.619 & 0.376 & 0.004 & 0.001 & 0.000 & 0.000 \\
& 300B & 500B & 0.548 & 0.444 & 0.004 & 0.003 & 0.002 & 0.000 \\
& 400B & 500B & 0.496 & 0.492 & 0.007 & 0.003 & 0.002 & 0.000 \\
& 500B & 500B & 0.496 & 0.492 & 0.007 & 0.003 & 0.002 & 0.000 \\
& 600B & 500B & 0.491 & 0.487 & 0.017 & 0.005 & 0.000 & 0.000 \\
& 700B & 500B & 0.439 & 0.430 & 0.130 & 0.001 & 0.000 & 0.000 \\
& 800B & 500B & 0.439 & 0.430 & 0.130 & 0.001 & 0.000 & 0.000 \\
& 900B & 500B & 0.404 & 0.403 & 0.183 & 0.006 & 0.003 & 0.000 \\
& 1000B & 500B & 0.395 & 0.387 & 0.214 & 0.003 & 0.001 & 0.000 \\
\hline
\multirow{9}{*}{1.8B} & 200B & 500B & 0.825 & 0.165 & 0.005 & 0.004 & 0.001 & 0.000 \\
& 300B & 500B & 0.619 & 0.376 & 0.004 & 0.001 & 0.000 & 0.000 \\
& 400B & 500B & 0.548 & 0.444 & 0.004 & 0.003 & 0.002 & 0.000 \\
& 500B & 500B & 0.548 & 0.444 & 0.004 & 0.003 & 0.002 & 0.000 \\
& 600B & 500B & 0.496 & 0.492 & 0.007 & 0.003 & 0.002 & 0.000 \\
& 700B & 500B & 0.496 & 0.492 & 0.007 & 0.003 & 0.002 & 0.000 \\
& 800B & 500B & 0.496 & 0.492 & 0.007 & 0.003 & 0.002 & 0.000 \\
& 900B & 500B & 0.491 & 0.487 & 0.017 & 0.005 & 0.000 & 0.000 \\
& 1000B & 500B & 0.491 & 0.487 & 0.017 & 0.005 & 0.000 & 0.000 \\
\hline
\multirow{9}{*}{1.2B} & 200B & 500B & 0.926 & 0.066 & 0.006 & 0.002 & 0.000 & 0.000 \\
& 300B & 500B & 0.758 & 0.229 & 0.012 & 0.001 & 0.000 & 0.000 \\
& 400B & 500B & 0.619 & 0.376 & 0.004 & 0.001 & 0.000 & 0.000 \\
& 500B & 500B & 0.619 & 0.376 & 0.004 & 0.001 & 0.000 & 0.000 \\
& 600B & 500B & 0.548 & 0.444 & 0.004 & 0.003 & 0.002 & 0.000 \\
& 700B & 500B & 0.496 & 0.492 & 0.007 & 0.003 & 0.002 & 0.000 \\
& 800B & 500B & 0.496 & 0.492 & 0.007 & 0.003 & 0.002 & 0.000 \\
& 900B & 500B & 0.496 & 0.492 & 0.007 & 0.003 & 0.002 & 0.000 \\
& 1000B & 500B & 0.496 & 0.492 & 0.007 & 0.003 & 0.002 & 0.000 \\
\hline
\end{tabular}
\end{table}

\section{Generalization to Refinedweb}
\label{app:generalization}

To evaluate the robustness and generalization capability of the InfoLaw across different data distributions, we conducted an additional series of verification experiments on the RefinedWeb dataset \citep{RefinedWeb}.

\textbf{Experimental Setup.} 
We followed the identical data preprocessing, LayerMix sampling, and training procedures described in Section \ref{sub:layermix} and Section \ref{sec:Training-setup}, with the sole exception of replacing the source corpus with RefinedWeb. Due to time and computational constraints, we limited the scope of this study to three LayerMix sampling configurations: HQ (High Quality) and LQ (Low Quality) were used for parameter fitting (interpolation), while MLQ (Medium-Low Quality) was held out for extrapolation testing. For each configuration, we trained models at three specific scales: 302M, 566M, and 1.2B parameters.

\textbf{Fitting and Extrapolation.} 
We applied the fitting methodology outlined in Section~5.2. Our analysis yielded two key observations:
\begin{itemize}
    \item \textbf{Consistency of Quality Density ($f$):} The fitted values for the quality density function $f_d$ were numerically very close to those derived from our primary dataset. Specifically, the fitted parameter $\theta$ is 0.93 for RefinedWeb, which is remarkably close to the value of 0.92 obtained from our primary dataset. We attribute this similarity to the fact that RefinedWeb \citep{RefinedWeb} is also derived from Common Crawl \citep{commoncrawl}; despite employing different filtering strategies, the shared underlying data source results in a comparable information density distribution.
    \item \textbf{Optimization of $\lambda(N)$:} In the main experiments, we modeled the relationship between the parameter $\lambda(N)$ and model scale $N$ using a logarithmic curve. However, due to the limited number of data points in this verification set (only three distinct model scales), fitting a robust $\lambda(N) - N$ curve was not feasible. Consequently, we skipped the curve fitting step for $\lambda(N)$ and directly searched for the optimal $\lambda$ values corresponding to the specific model sizes (302M, 566M, and 1.2B).
\end{itemize}

\begin{figure}
    \centering
    \includegraphics[width=0.9\textwidth]{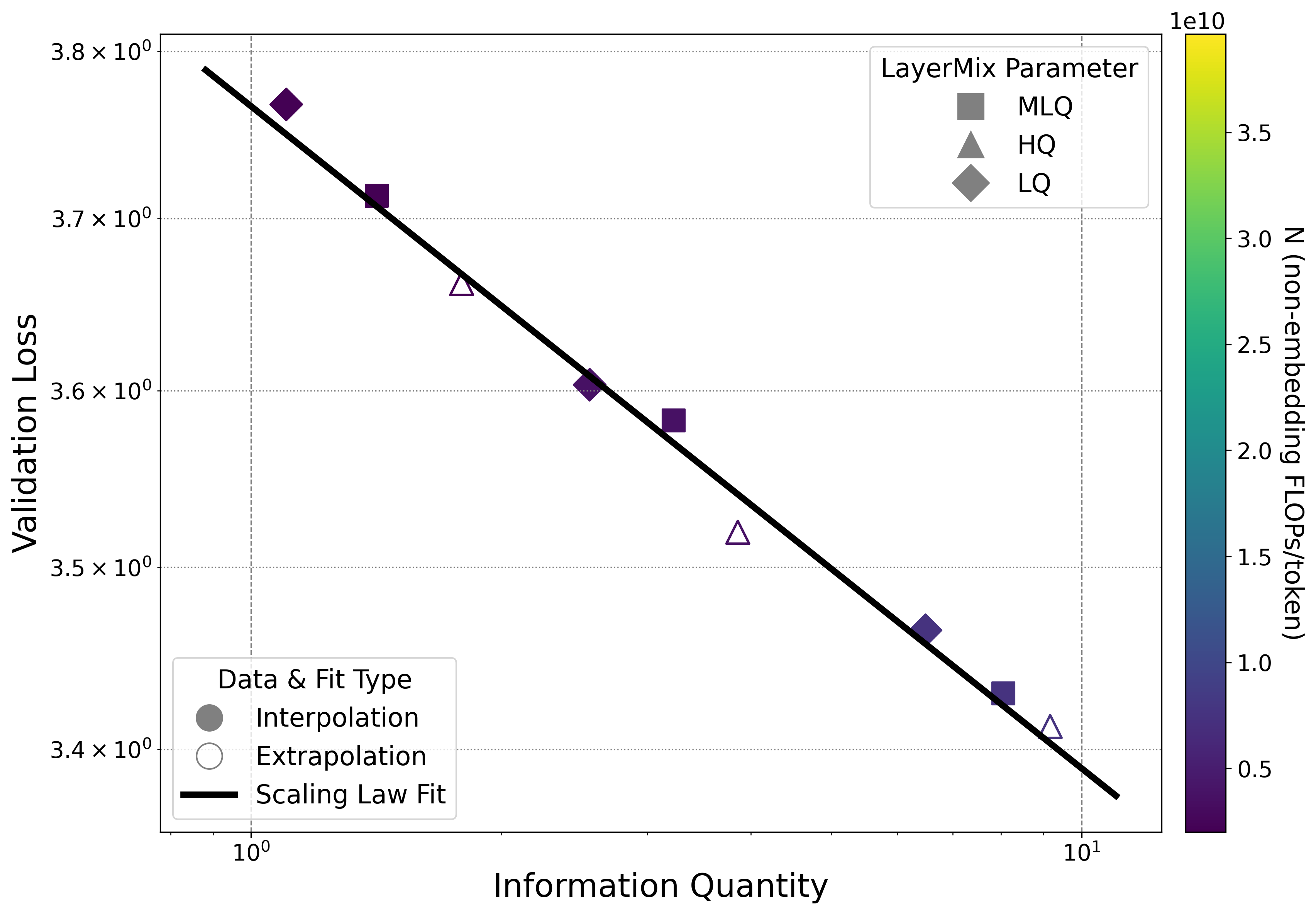}
    \caption{The Unified Information-Loss Scaling Law on the RefinedWeb dataset.}
    \label{fig:refinedweb_info_scaling_all}
\end{figure}

\textbf{Results.} 
Using the parameters fitted on the HQ and LQ configurations, we predicted the validation loss for the unseen MLQ configuration, as illustrated in Figure \ref{fig:refinedweb_info_scaling_all}. The InfoLaw demonstrated strong predictive accuracy on the RefinedWeb dataset, achieving a maximum absolute error of 0.36\% and a mean absolute percentage Error 0.24\% on the extrapolated MLQ experiments. These results further corroborate that the InfoLaw effectively captures the fundamental trade-offs between data quality, repetition, and compute scale, independent of the specific underlying data source.

\begin{figure}
    \centering
    \includegraphics[width=0.6\textwidth]{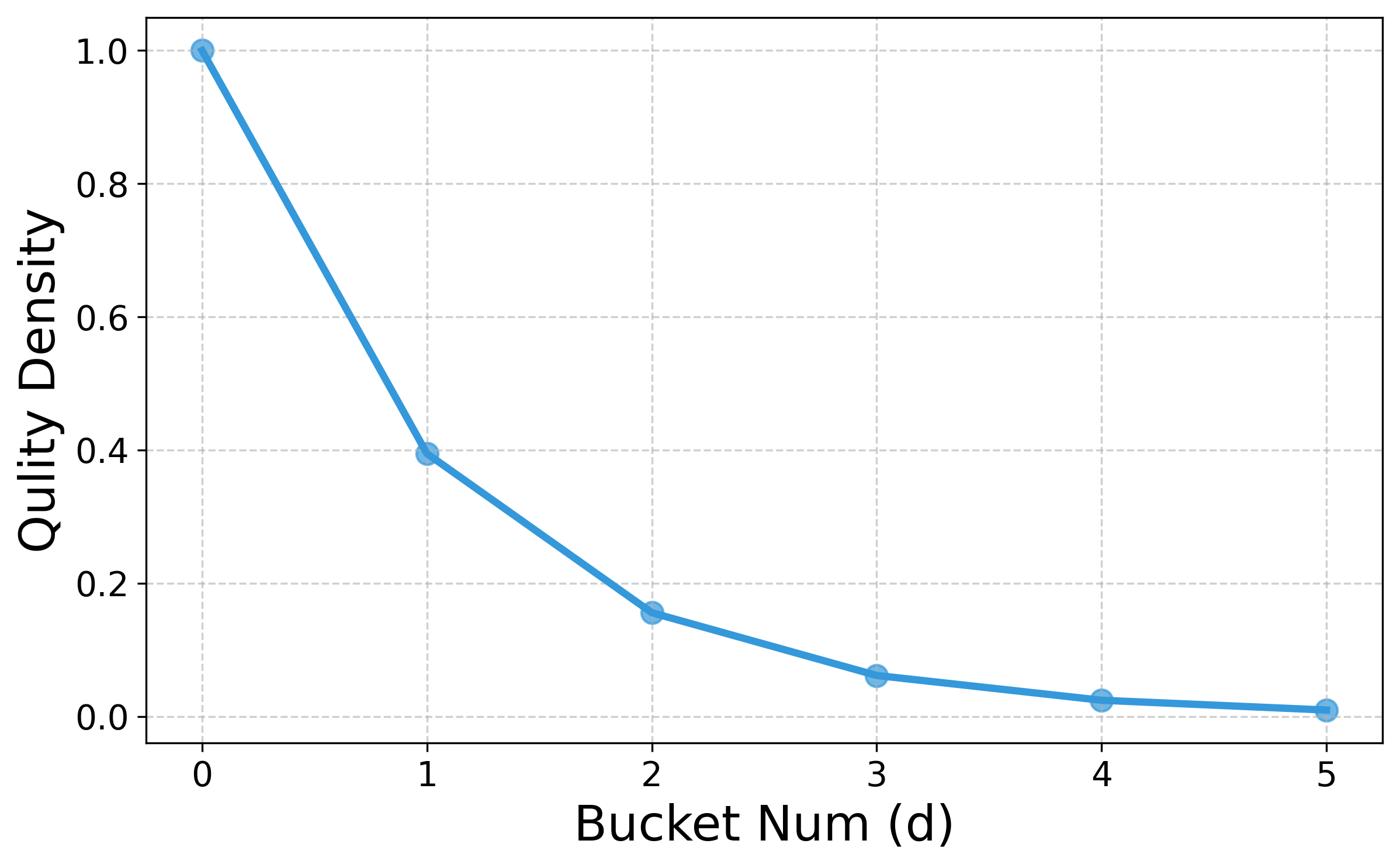}
    \caption{The fitted quality density function $f_d$ on the RefinedWeb dataset.}
    \label{fig:refinedweb_f}
\end{figure}

\section{Limitation}

We note several limitations of our work. Our data bucketing is based on a fixed, empirical heuristic. We have not performed ablation studies to determine the optimal number or boundaries of these quality tiers. A more systematic approach to data partitioning could further improve the model's predictive accuracy. And while we observe that the overtrain degree $m$ systematically shifts the scaling law curve, a theoretical explanation for this behavior is still needed. These areas present clear avenues for future work.

\end{document}